\PassOptionsToPackage{margin=1in}{geometry}
\documentclass[multiauthors,onecolumn]{LaTeXclass/cohere}

\usepackage{ragged2e}
\usepackage{lipsum}
\usepackage{pdfpages}
\usepackage{colortbl}
\usepackage{tabulary}
\usepackage{longtable}
\usepackage{array}
\usepackage{placeins}
\usepackage{tablefootnote}
\usepackage{xcolor}
\usepackage{tcolorbox}
\usepackage{wrapfig}
\usepackage{breqn} 

\usepackage{bm}
\usepackage{graphicx}
\usepackage{booktabs}
\usepackage{microtype}
\usepackage{hyperref}

\usepackage{pifont}
\definecolor{deeppurple}{HTML}{9e02f7}
\definecolor{forestgreen}{HTML}{2e7d43}
\usepackage{multirow}
\usepackage{tabularx}

\usepackage{amsmath,amssymb,amsfonts,mathtools,amsthm}
\usepackage[margin=1in]{geometry}

\usepackage{fancyvrb}
\usepackage{fvextra}
\usepackage{comment}

\usepackage[utf8]{inputenc}

\usepackage{arydshln}

\usepackage{xspace} 
\usepackage{makecell} 
\usepackage{multirow}

\usepackage{nicematrix}
\usepackage{comment}

\usepackage{xcolor}
\definecolor{fixedbestblue}{RGB}{30,90,160}

\setcitestyle{number,square}

\title{CALIBER: Calibrating Confidence Before and After Reasoning in Language Models}

\author{name={Conor Finlay\fa},affiliation={1}}
\author{name={Joshua Kurien\fa},affiliation={1}}
\author{name={Saurabh Dash},affiliation={1}}
\author{name={Marzieh Fadaee\psa},affiliation={1}}
\author{name={Beyza Ermis\psa},affiliation={1}}
\affiliations{
    \item[1] Cohere Labs
}

\corresponding[*]{\{\url{beyza}, \url{marzieh}\}\url{{@cohere.com}}}

\abstract{
\justifying

Reasoning language models are increasingly asked not only to answer difficult questions, but also to estimate their likelihood of success. Existing methods typically elicit confidence only once: either before thinking or after answering. We argue that confidence in reasoning models is state-dependent: before thinking, confidence should estimate the chance of the model correctly solving the prompt, while after thinking it should predict whether the realized answer is likely to be correct. This distinction determines the appropriate supervision target: prompt-level success should supervise confidence estimates made after seeing the prompt, while individual answer-level correctness should supervise confidence estimates made after answering.
We introduce \textsc{CALIBER} (Calibration Before and After Reasoning), which elicits both estimates and supervises each with the target matched to its information state. 
Under this unified protocol, \textsc{CALIBER} reduces Expected Calibration Error (ECE) by 52.5\% over the strongest single-confidence baseline on BigMathDigits for the 7B model, while achieving the best Brier score and AUROC, and remains within 2.1 points of the best accuracy. Further, on a larger 30B model, \textsc{CALIBER} achieves the best ECE on BigMathDigits while remaining competitive in Brier score and AUROC. Out of distribution, it achieves the best ECE and Brier score on GPQA and TriviaQA, and remains competitive on SimpleQA. Ablations further show that this position–target alignment is most beneficial under distribution shift where it consistently reduces calibration error across all out-of-distribution benchmarks.
}

\renewcommand{\FONTauthors}{\bfseries\Large}
\renewcommand{\FONTauthor}{\bfseries\Large}
\begin{document}

\section{Introduction}
\label{introduction}
Large language models increasingly solve complex tasks through extended thinking. In many applications, a model needs not only to answer, but also to reliably estimate whether its answer is correct. A well-calibrated model signals to users how trustworthy its answers are, supports verification of low-confidence answers, and can be deployed more safely in settings where incorrect answers are costly \citep{Geifman2017Selective, OnCalibration, TamingOverconfidence}.

Prior LLM calibration work explores several confidence elicitation strategies, including token probabilities, consistency across sampled responses, and explicit verbalized confidence \citep{LLMsKnowWhatTheyDontKnow, lin2022teaching, SemanticUncertainty, CanLLMsExpressUncertainty}. We focus on verbalized confidence, where the model reports an explicit probability of correctness as part of its response, since such estimates can be easily extracted from reasoning models \citep{SaySelf, ConfTuner, UncertaintyDistillation}.

Reasoning models expose additional information beyond a single answer by including thinking traces alongside their answers, but standard calibration methods are typically designed for single-answer outputs. These extended thinking traces are produced before the model commits to a final response, so the model's information state changes substantially between the start and end of generation. For reasoning models, the relevant probability concerns the correctness of a solution produced through a multi-step reasoning process. Thus, confidence before and after thinking estimate distinct quantities: the former estimates whether the current policy is likely to solve the prompt, while the latter estimates whether the realized answer is correct, taking into account the reasoning process that produced it. Because these estimates are produced with access to different information, they require different supervision targets.

Recent methods address calibration in reasoning models through reinforcement learning with verifiable rewards (RLVR) \citep{DeepseekGRPO, AllenInstituteRLVR}. Methods vary along two axes: the \emph{position} of the confidence estimate, either before thinking \citep{coca} or after thinking \citep{rlcr, corea}; and the \emph{calibration target}, either instance-wise correctness \citep{rlcr}, group-wise success across rollouts of the same prompt \citep{corea, coca}, or a hybrid of the two \citep{DCPO}.
Existing methods choose one position and one target, but do not ask whether that pairing is optimal given the information state at which confidence is elicited.

We argue that the appropriate supervision target is determined by elicitation position: pre-thinking confidence, which conditions only on the prompt, should be supervised against the prompt-level success rate, while post-thinking confidence, which additionally conditions on the realized trace and answer, should be supervised against the correctness of that realized response. Conflating these introduces a mismatch between what the model is asked to estimate and what it is trained to predict.

We introduce \textsc{CALIBER}, a framework that elicits confidence both before and after reasoning and supervises each estimate with the calibration target matched to its information state. Rather than forcing a single estimate to serve two roles, the framework lets each one specialize: pre-confidence captures a model's ability to solve the question, while post-confidence captures whether its stated answer is correct. The gap between them provides a per-example diagnostic of how thinking changes the model's uncertainty.

Isolating the contribution of position–target alignment requires a controlled setting, since prior methods differ simultaneously in confidence position, calibration target, credit assignment technique, and reinforcement learning (RL) algorithm. We therefore compare a controlled set of prior-inspired confidence-position and supervision-target pairings under a shared training and evaluation protocol. Across mathematical reasoning and question-answering benchmarks, CALIBER improves calibration without substantially sacrificing task performance. 
On BigMathDigits, CALIBER reduces ECE over the strongest single-confidence baseline by 52.5\% for the 7B model and achieves the best ECE for the 30B model. These gains transfer out of distribution, where CALIBER obtains the best ECE and Brier score on GPQA and TriviaQA while remaining competitive on SimpleQA.

\noindent Our contributions are:
\begin{itemize}
\item We frame confidence in reasoning models as state-dependent: pre-thinking and post-thinking confidence correspond to distinct uncertainty quantities, question solvability and answer-level correctness, and naturally align with group-wise and instance-wise calibration targets respectively.
\item We introduce \textsc{CALIBER}, which emits both pre- and post-confidence estimates and supervises each with the supervision target matched to its information state.
\item We present a controlled comparison against prior calibration strategies under a shared protocol, isolating the contribution of position--target alignment from confounding design choices.
\item We analyze confidence updates by comparing pre- and post-thinking confidence as a per-example signal of how the model's uncertainty evolves with thinking.
\end{itemize}

\section{Related Work}
\label{sec:related_work}

\textbf{Confidence estimation for LLMs.}
Confidence estimation techniques for LLMs include token probabilities, sampling-based consistency, internal-state probes, and explicit verbalized confidence~\citep{geng2024survey}. Token probabilities are easy to obtain, but the estimates are over token-level predictions rather than answer-level correctness \citep{CanLLMsExpressUncertainty}. Verbalized confidence instead asks the model to report a scalar estimate of the probability that its answer is correct. Prior work shows that models can be trained to express uncertainty in words \citep{lin2022teaching}, with later methods improving confidence estimates through supervised fine-tuning, self-reflective rationales, confidence tuning, or distillation \citep{kapoor2024large,SaySelf,ConfTuner,UncertaintyDistillation}. Other work suggests that models may encode information about their own correctness internally, even when this information is not reliably surfaced in their responses \citep{LLMsKnowWhatTheyDontKnow,DeepmindVerbalConfidence}.

\textbf{RLVR for calibrated verbalized confidence.}
A current trend for training calibrated reasoning models is to modify standard correctness objectives in RLVR to include calibration rewards \citep{DeepseekGRPO,AllenInstituteRLVR,bani2025rewarding}. Reinforcement Learning with Calibration Rewards (RLCR) augments the standard RLVR correctness reward with a Brier-score reward that encourages the model's verbalized confidence for a rollout to match that rollout's binary correctness \citep{rlcr}. COllaborative REAsoner (COREA) instead trains confidence to match the average correctness of a group of rollouts from the same prompt \citep{corea}. Decoupled Calibration Policy Optimization (DCPO) adopts a hybrid approach, interpolating between prompt-level group accuracy and individual rollout correctness \citep{DCPO}. While most methods elicit confidence estimates after giving an answer, Co-optimized Confidence and Answers (CoCA) elicits confidence before the model produces its answer \citep{coca}. Both DCPO and CoCA utilize segmented credit assignment to compute and apply correctness and calibration advantages exclusively to the tokens that contribute to the respective rewards.

Prior RLVR methods vary in where confidence is elicited and how it is supervised \citep{rlcr,corea,coca,DCPO}. However, these choices have largely been studied in single-confidence settings. RLCR, COREA, and DCPO elicit confidence after answering, but require a single post-answer estimate to capture different forms of uncertainty. CoCA elicits confidence before answering, but does not account for how the realized reasoning trace and answer should update that estimate.
The complementary setting where both pre- and post-thinking confidences are elicited, and each is matched to the target appropriate given its information state, has not been systematically studied. \textsc{CALIBER} addresses this gap.

\section{Preliminaries}
\label{sec:preliminaries}

This section introduces the notation and evaluation criteria used throughout the paper. We first define the calibration metrics used to evaluate confidence estimates, then describe how the position of a verbalized confidence estimate determines what information it can condition on, and finally formalize the calibration targets used in RLVR-style training.

\subsection{Calibration Metrics}
\label{sec:calibration}
Let $p_i \in [0,1]$ denote a model's predicted probability that its answer to example $i$ is correct, and let $c_i \in \{0,1\}$ denote whether the answer is correct. A confidence estimator is calibrated if predictions assigned confidence $p$ are correct with probability $p$: 
\begin{equation}
    \Pr(c=1 \mid p_i=p) = p.
\end{equation}
In practice, calibration is evaluated through complementary metrics. We distinguish \emph{absolute calibration}, which measures agreement between confidence and empirical accuracy, from \emph{failure prediction}, which measures whether confidence distinguishes correct from incorrect answers \citep{CanLLMsExpressUncertainty}.

\textbf{Expected Calibration Error.}
Expected Calibration Error (ECE) measures absolute calibration by partitioning predictions into $M$ confidence bins $B_1,\ldots,B_M$ and computing
\begin{equation}
    \mathrm{ECE} =
    \sum_{m=1}^{M} \frac{|B_m|}{n}
    \left|\mathrm{acc}(B_m)-\mathrm{conf}(B_m)\right|.
\end{equation}
ECE measures whether confidence values match empirical accuracy \citep{OnCalibration}, but can be misleading on its own: a model that predicts its dataset-average accuracy for every example can appear well calibrated while providing no information about which answers are likely to be wrong.

\textbf{Brier score.}
The Brier score measures the squared error between confidence and correctness:
\begin{equation}
    \mathrm{Brier} =
    \frac{1}{n}\sum_{i=1}^{n}(p_i-c_i)^2.
\end{equation}
Lower Brier scores indicate better probabilistic predictions.

\textbf{AUROC.}
Treating confidence as a score for answer correctness, AUROC measures whether correct answers tend to receive higher confidence than incorrect ones. We report AUROC together with ECE and Brier because a confidence estimator can be well calibrated on average while still failing to discriminate successes from failures. We use this as our failure prediction metric.

\subsection{Verbalized Confidence for Reasoning LLMs}
\label{sec:verbalized_conf}
Verbalized confidence methods ask a model to output an explicit probability of correctness together with its response. For reasoning models, we treat the response as a structured sequence containing a thinking trace, an answer, and one or more confidence estimates. A standard post-answer confidence format is:
\begin{verbatim}
<thinking> ... </thinking>
<answer> ... </answer>
<confidence> ... </confidence>
\end{verbatim}
In this format, the confidence estimate is emitted after the model has already produced both its thinking trace and final answer.

The location of the confidence estimate determines what information it can condition on. Let $x$ be the prompt, $z$ the thinking trace, $\hat{y}$ the model's answer, and $c \in \{0,1\}$ the answer's correctness. A confidence estimate emitted before thinking is question-conditioned:
\begin{equation}
    p_{\mathrm{question}} = P(c=1 \mid x).
\end{equation}
where the probability is taken over responses sampled from the current policy $\pi_\theta$.
It estimates the probability that the model will solve the prompt before a particular reasoning trace or answer is observed. By contrast, a confidence estimate emitted after thinking and answering is answer-conditioned:
\begin{equation}
    p_{\mathrm{answer}} = P(c=1 \mid x, z, \hat{y}).
\end{equation}
It estimates whether the model's realized answer is correct.

This distinction is central to our method. Pre-confidence and post-confidence are not two copies of the same scalar: they correspond to different information states. The pre-confidence should be trained with a target that reflects prompt-level solvability, while the post-confidence should be trained with a target that reflects answer-level correctness.

\subsection{RLVR, Rollouts, and Calibration Targets}
\label{sec:targets}
RLVR trains models on tasks where responses can be checked automatically, such as mathematical reasoning \citep{AllenInstituteRLVR}. In GRPO-style training, for each prompt $x$, the current policy samples $N$ responses, or \emph{rollouts}, and computes rewards for each rollout relative to the group \citep{DeepseekGRPO}. Within this setup, rollouts can be scored with calibration rewards to improve confidence estimation behavior.

For a prompt $x$, let $c_i \in \{0,1\}$ indicate whether rollout $i$ is correct. Calibration rewards compare a predicted confidence $p_i$ to a target $t_i$, typically using a squared-error reward of the form
\begin{equation}
    R_{\mathrm{cal},i} = - (p_i - t_i)^2.
\end{equation}
The target $t_i$ can be defined in several ways.

\textbf{Instance-level target.}
The instance-level target is the correctness of the individual rollout:
\begin{equation}
    t_i^{\mathrm{inst}} = c_i.
\end{equation}
This target is appropriate for answer-conditioned confidence because it supervises whether the particular sampled answer is correct.

\textbf{Group-level target.}
The group-level target is the empirical accuracy of all $N$ rollouts for the same prompt:
\begin{equation}
    t^{\mathrm{group}}(x) =
    \frac{1}{N}\sum_{j=1}^{N} c_j.
\end{equation}
This target is appropriate for question-conditioned confidence because it estimates the probability that the current policy solves the prompt before a particular answer is observed.

\textbf{Hybrid target.}
Hybrid targets interpolate between instance-level and group-level supervision:
\begin{equation}
    t_i^{\mathrm{hybrid}} =
    \lambda c_i + (1-\lambda)\frac{1}{N}\sum_{j=1}^{N} c_j.
\end{equation}
Such targets combine prompt-level and answer-level information in a single scalar \citep{DCPO}.

\section{Method}
\label{sec:method}

We introduce \textsc{CALIBER}, a framework for training reasoning models to emit two calibrated verbalized confidence estimates in a single response, each supervised with a target corresponding to the information state at which the estimate is emitted. The model first emits a \emph{pre-confidence} estimate before thinking, then produces a thinking trace and answer, and finally emits a \emph{post-confidence} estimate. Pre-confidence is supervised with a group-wise target that reflects expected prompt-level success rate, while post-confidence is supervised with an instance-wise target that reflects the correctness of the realized answer.

\subsection{State-Dependent Calibration}
\label{sec:state_dep_cal}

\textsc{CALIBER}'s key principle is that confidence estimates depend on the model's information state. The appropriate supervision target therefore depends on where an estimate is elicited within the response. Before thinking, the model has not yet produced a trace or answer, so its confidence cannot depend on the quality of a realized solution, but should instead estimate the expected success of the current policy on the prompt. After thinking, the model has produced a specific trace and answer, so its confidence can condition on the realized response and should estimate whether that particular rollout is correct. A target that does not match the confidence elicitation position therefore creates a mismatch between what the model can estimate and what it is trained to predict. For example, using an instance-wise target for pre-confidence is mismatched because it trains a prompt-level estimate with a rollout-level reward.


\subsection{Setup and Notation}
\label{sec:notation}
We use RLVR training where, for each prompt $x$, we sample $N$ rollouts from the current policy:
\begin{equation}
    \{r_i\}_{i=1}^{N} \sim \pi_\theta(\cdot \mid x).
\end{equation}
Each rollout $r_i$ contains a pre-confidence estimate $p_{\mathrm{pre},i}$, a thinking trace $z_i$, an answer $\hat{y}_i$, and a post-confidence estimate $p_{\mathrm{post},i}$. Let $y^\star$ denote the reference answer and let $V(\hat{y},y^\star)\in\{0,1\}$ denote the task verifier. 
The correctness of a rollout is a binary value defined as
\begin{equation}
c_i = V(\hat{y}_i, y^\star),
\end{equation}
and the empirical success rate of the current policy on prompt $x$ is estimated as
\begin{equation}
    \hat{p}(x) = \frac{1}{N}\sum_{i=1}^{N} c_i.
\vspace{-1mm}
\end{equation}
This group-wise quantity is shared by all rollouts for the same prompt. Because $\hat{p}(x)$ is computed from sampled rollouts, it estimates the current policy's empirical success probability on $x$, rather than the prompt's intrinsic difficulty.

\subsection{Output Format}

The model is prompted to produce a structured response with confidence estimates wrapped in special tags. For example:
\begin{verbatim}
<pre_confidence>p_pre</pre_confidence>
<thinking> ... </thinking>
<answer> ... </answer>
<post_confidence>p_post</post_confidence>
\end{verbatim}
The pre-confidence estimate is emitted before thinking, so it can condition only on prompt $x$. The post-confidence estimate is emitted after the thinking trace $z_i$ and answer $\hat{y}_i$, so it conditions on the realized response. This implements the state-dependent calibration discussed in Section~\ref{sec:state_dep_cal}: pre-confidence estimates prompt-level solvability, while post-confidence estimates answer-level correctness.

To ensure that the model follows the required format, we include a binary format reward $R_{\mathrm{fmt},i}$ that equals 1 only when all required tags are present, the answer can be extracted, and the confidence values are valid numbers in $[0,1]$. All other rewards are applied only when the rollout is well-formatted.

\subsection{Calibration Targets}

Each confidence estimate is assigned the calibration target that matches its position in the response.

\textbf{Pre-confidence target.}
The pre-confidence estimate is trained with the group-wise success rate target:
\begin{equation}
    t_i^{\mathrm{pre}} = \hat{p}(x)
    = \frac{1}{N}\sum_{j=1}^{N} c_j.
\end{equation}
Since this target averages over all rollouts in the group, it is not conditioned on any particular answer. It therefore estimates the current policy's success probability on prompt $x$, rather than the correctness of any particular rollout. This makes it the appropriate target for pre-confidence: before thinking, the model should estimate its expected success under the policy it is about to execute. Correct and incorrect rollouts sampled from the same prompt therefore receive the same pre-confidence target, so the target supervises prompt-level solvability rather than the correctness of a specific generated answer.

\textbf{Post-confidence target.}
The post-confidence estimate is trained with the instance-wise correctness target:
\begin{equation}
    t_i^{\mathrm{post}} = c_i.
\end{equation}
This target supervises whether rollout $i$ produced the correct answer, and is therefore appropriate for post-confidence, which can condition on the realized thinking trace and generated answer.

Two consequences of this assignment are worth noting. First, the two confidence scores are not constrained to agree: $p_{\mathrm{post},i}$ may rise above $p_{\mathrm{pre},i}$ when the model has evidence that its solution is reliable, or fall below it when the generated solution appears uncertain or incorrect. Second, the gap $p_{\mathrm{post},i} - p_{\mathrm{pre},i}$ is itself a quantity of interest, since it summarizes how the model's expressed uncertainty changes after thinking.

\subsection{Training Objective}
Our training objective enables the two confidence estimates to specialize for their complementary roles. It consists of the format reward, the group-wise pre-confidence reward, the instance-wise post-confidence reward, and the binary correctness reward for the model's answer. All reward components are equally weighted to keep the comparison controlled. The reward for rollout $i$ is
\begin{equation}
    R_i =
    R_{\mathrm{acc},i}
    + R_{\mathrm{fmt},i}
    + R_{\mathrm{grp},i}^{\mathrm{pre}}
    + R_{\mathrm{inst},i}^{\mathrm{post}}
    \label{eq:total-reward}
\end{equation}
The terms denote binary answer correctness
$R_{\mathrm{acc},i}=c_i$,\allowbreak{}
format validity
$R_{\mathrm{fmt},i}=\mathbf{1}[\text{response is well}\allowbreak{}\text{formatted}]$,
pre-confidence calibration
$R_{\mathrm{grp},i}^{\mathrm{pre}}
=1-\left(p_{\mathrm{pre},i}-\hat{p}(x)\right)^2$,\allowbreak{}
and post-confidence calibration
$R_{\mathrm{inst},i}^{\mathrm{post}}
=1-\left(p_{\mathrm{post},i}-c_i\right)^2$.

We retain an accuracy reward because training only for calibration can admit degenerate solutions: a model can produce incorrect answers with low confidence and still receive a high calibration reward \citep{coca,rlcr}. 
The accuracy reward also ensures that calibration gains are not obtained by sacrificing task performance, since the calibrated model should remain competitive with one trained using standard correctness-only RLVR.

\section{Experimental Setup}
\label{sec:experimental_setup}

\textbf{Models and Training.}
We run experiments using a 7B-parameter model and a 30B-parameter model, each trained for 500 calibration RL steps. The 7B model is based on the Command R7B architecture\footnote{\href{https://huggingface.co/CohereLabs/c4ai-command-r7b-12-2024}{https://huggingface.co/CohereLabs/c4ai-command-r7b-12-2024}}, and the 30B model is based on the North Mini Code architecture\footnote{\href{https://huggingface.co/CohereLabs/North-Mini-Code-1.0}{https://huggingface.co/CohereLabs/North-Mini-Code-1.0}}.

Before calibration training, all methods are trained with a mix of SFT and RL using only the format reward, so that the model reliably includes the special tags for the confidence estimates. All calibration experiments then start from this shared format-learned checkpoint and use the same GRPO-based algorithm with no standard deviation normalization, with no KL penalty, the same number of rollouts per prompt, and the same reward filters. Additional training and evaluation details are provided in Appendix~\ref{app:training_details}.

\textbf{Training \textsc{CALIBER}.}
We train \textsc{CALIBER} in two stages. We start with a \textit{warmup} stage where we optimize only the pre-confidence group-wise calibration reward for a fixed number of steps. In the second stage, we use the remaining calibration steps to jointly train with the pre-confidence group-wise reward, the post-confidence instance-wise reward, and the correctness reward. The format reward is present in both stages.

This staged schedule addresses an optimization difficulty that arises in this multi-reward setting when the pre-confidence reward is combined with the other rewards from the start. Under GRPO, the rewards for each rollout are summed and converted into group-relative advantages, so an individual reward's influence depends on how much it varies across a prompt's rollouts. The group-wise target $\hat{p}(x)$ is shared across a prompt's rollouts, so if the model's pre-confidence estimates vary little across rollouts, so do the corresponding pre-confidence rewards. In contrast, the post-confidence and correctness rewards are tied to per-rollout binary correctness and vary much more. As a result, pre-confidence receives a weak training signal, and the estimates collapse to a handful of values early in training and never recover. This issue is both specific to the group-wise nature of the pre-confidence reward and a more general characteristic of multi-reward RL, where higher-variance rewards can dominate.

Training the pre-confidence reward first encourages enough diversity in pre-confidence estimates for all the individual rewards to train effectively in the second stage. 
We use fixed warmup lengths of 400 steps for the 7B model and 100 steps for the 30B model, selected because the range of pre-confidence estimates was consistently high by these points.

\begin{figure}
    \centering
    \includegraphics[width=0.6\linewidth]{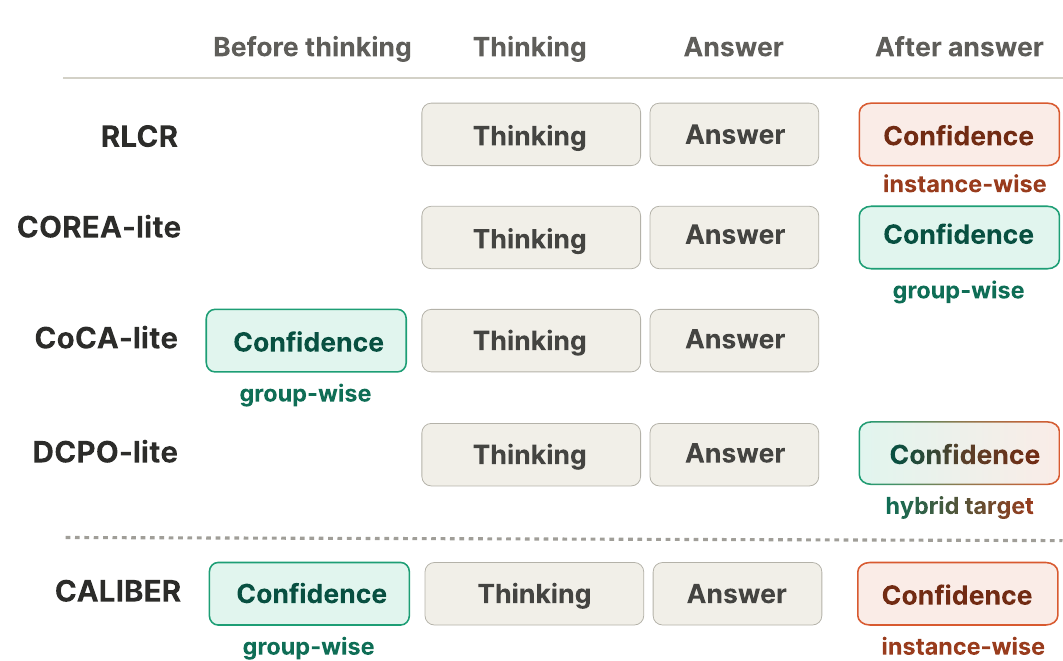}
    \caption{Confidence positions and calibration targets for the evaluated methods.}
    \label{fig:confidence_methods}
\vspace{-2mm}
\end{figure}

\textbf{Baselines.}
We compare \textsc{CALIBER} against baselines that isolate two design choices: confidence position and calibration target. To keep the comparison controlled, no method uses segmented credit assignment, where correctness and calibration advantages are applied to different token subsets, and all calibration rewards use squared error between predicted confidence and target. Methods labeled \emph{-lite} are simplified adaptations of prior work under this shared setup: they preserve the original confidence position and target choice, while omitting method-specific details that would confound comparison.
Figure~\ref{fig:confidence_methods} summarizes the confidence format and calibration target for each method. \textbf{Base} is the format-trained checkpoint before calibration RL. \textbf{RLVR} is trained with an answer-correctness reward but no explicit calibration reward. \textbf{RLCR} adds post-confidence instance-wise calibration with an accuracy reward \citep{rlcr}. \textbf{COREA-lite} uses post-confidence group-wise calibration with an accuracy reward \citep{corea}. \textbf{CoCA-lite} uses pre-confidence group-wise calibration with an accuracy reward \citep{coca}. \textbf{DCPO-lite} applies post-confidence calibration with a hybrid target that interpolates between group-wise and instance-wise supervision, alongside an accuracy reward. In our experiments, we set the interpolation weight to $\lambda = 0.5$, giving equal weight to both supervision types \citep{DCPO}.

\textbf{Datasets.}
All methods are trained on the same data: a filtered subset of Big-Math~\citep{BigMathDataset} from \citet{rlcr}. This dataset retains only problems with strictly numerical answers, enabling reliable verification, and is further filtered to include problems that a LLaMA-8B model solves between 0\% and 70\% of the time. This ensures diverse correctness signals for calibration training. We refer to the resulting 30{,}000-sample dataset as the BigMathDigits training set.

We evaluate in-distribution performance on the BigMathDigits evaluation split used by \citet{rlcr} which consists of 1000 samples. To assess out-of-distribution (OOD) generalization, we also evaluate on three question-answering datasets: TriviaQA~\citep{TriviaQA}, SimpleQA~\citep{SimpleQA}, and GPQA~\citep{GPQA}. For TriviaQA, we use the 1000-question subset from \citet{chhikara2025mind}; for SimpleQA, we use all 4326 samples and for GPQA, we use the 198-question Diamond subset.
For BigMathDigits, correctness is verified using heuristic and symbolic checks before falling back to an LLM-as-a-judge. GPQA is evaluated via regex matching against the correct multiple-choice letter, while SimpleQA and TriviaQA are evaluated using an LLM judge. For our LLM judge, we use Command A by ~\cite{cohere2025command}.

\textbf{Evaluation Protocol.}
We evaluate checkpoints every 100 steps over the full 500-step training run and select one checkpoint per method using the BigMathDigits evaluation split. Checkpoint selection uses the best combined rank across in-domain ECE and AUROC, with AUROC used to break ties. We report performance of this selected checkpoint on BigMathDigits as the in-domain result, and use the same checkpoint without further selection for all out-of-distribution evaluations. For methods with two confidence estimates, we evaluate the post-thinking confidence estimate.

For each question, we sample responses using a temperature of $0.3$ and top-p of $0.75$. We sample 3 responses per question for all datasets except GPQA, where we sample 12 responses with shuffled answer choices to cover different answer positions and reduce positional bias. Before computing metrics, we discard malformed generations, responses that do not follow the required confidence format, and responses with repetitive token loops.

\textbf{Metrics.}
We report \textbf{ECE} for absolute calibration, \textbf{Brier score} as a proper scoring rule, \textbf{AUROC} for failure prediction (whether correct rollouts receive higher confidence than incorrect ones), and \textbf{Accuracy} to verify that calibration is not obtained at the cost of correctness. We consider these metrics jointly because a model can be well calibrated on average while still failing to distinguish correct answers from incorrect ones \citep{CanLLMsExpressUncertainty}. A prototypical example of this is where the model always predicts its dataset-average accuracy. In this case, all estimates land in one confidence bin, yielding near-perfect ECE but very poor AUROC.

\section{Results}
\label{sec:results}

We evaluate \textsc{CALIBER} through comparisons designed to isolate the effects of confidence position, calibration target, and the two-confidence structure. These experiments address whether \textsc{CALIBER} improves calibration over pre-only or post-only confidence, whether the gains generalize out of distribution, whether the choice of supervision target for each estimate is important, and whether the model meaningfully updates its confidence after observing its own thinking trace and answer.

\begin{figure}[t]
    \centering
    \begin{subfigure}[t]{0.49\linewidth}
        \centering
        \includegraphics[width=\linewidth]{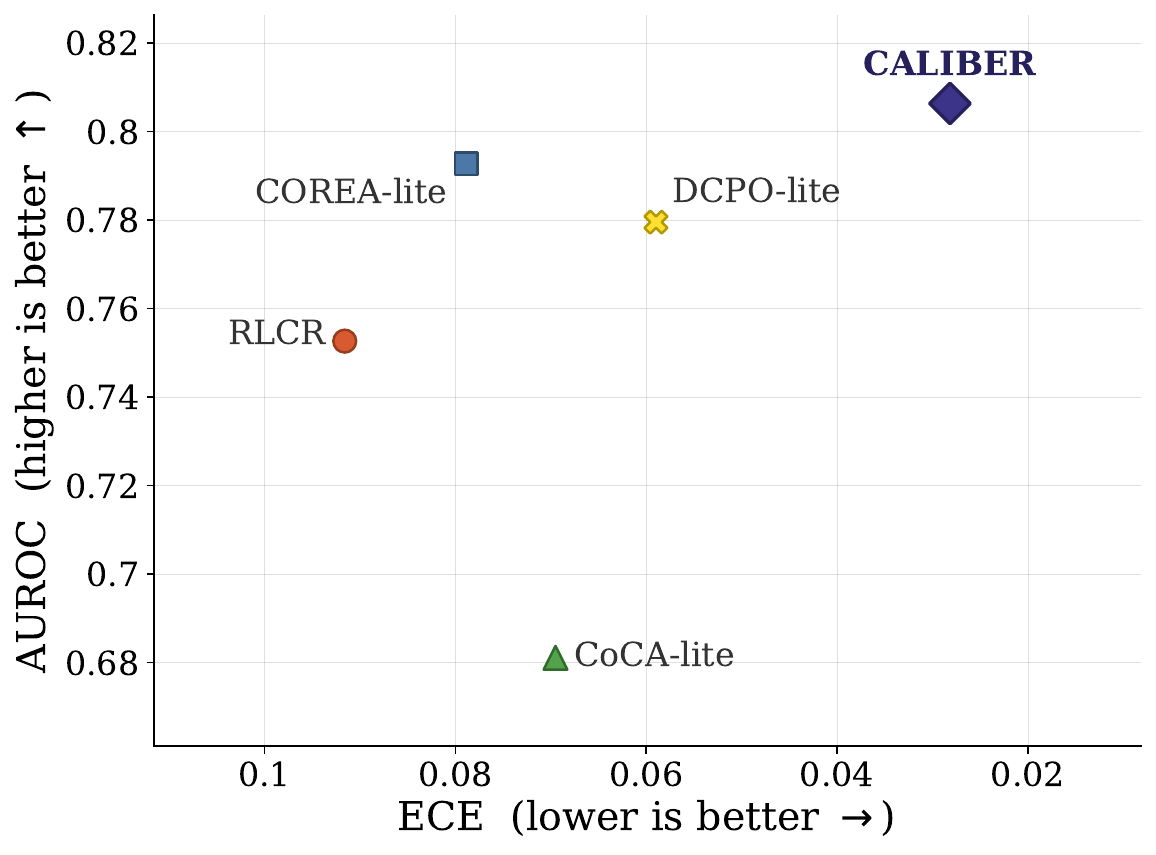}
        \caption{7B model.}
        \label{fig:bigmathdigits_pareto_7b}
    \end{subfigure}
    \hfill
    \begin{subfigure}[t]{0.49\linewidth}
        \centering
        \includegraphics[width=\linewidth]{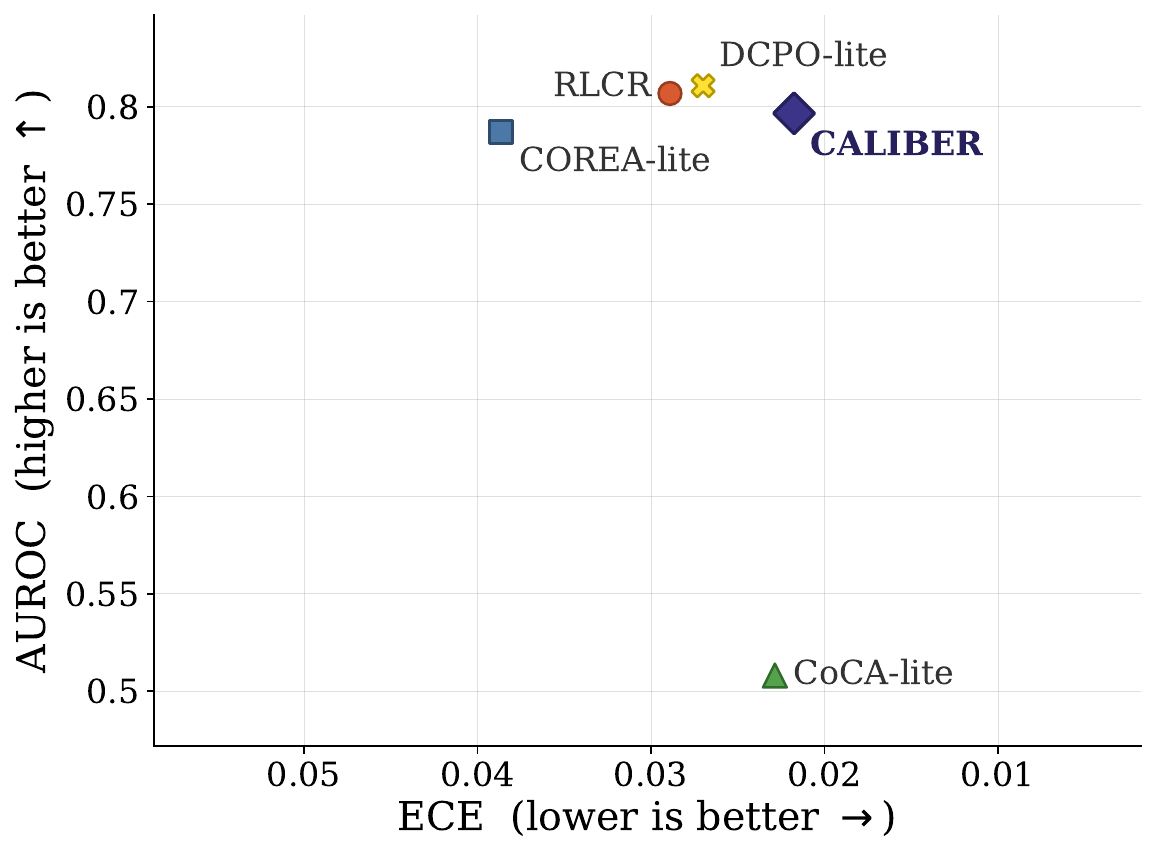}
        \caption{30B model.}
        \label{fig:bigmathdigits_pareto_30b}
    \end{subfigure}
    
    \caption{ECE--AUROC tradeoff on BigMathDigits. Lower ECE and higher AUROC are better. \textsc{CALIBER} gives the best tradeoff for the 7B model and remains on the low-ECE frontier for the 30B model.}
    \label{fig:BigMath_pareto_plot}
\end{figure}

\subsection{In-Domain Calibration on BigMathDigits}

We evaluate on the held-out BigMathDigits split to measure in-domain performance. Table~\ref{tab:bigmathdigits-selected-checkpoints} shows that \textsc{CALIBER} achieves the lowest ECE for both model sizes, indicating the strongest absolute calibration on BigMathDigits. For the 7B model, \textsc{CALIBER} also achieves the best Brier score and AUROC, improving all calibration metrics over the single-confidence baselines. At 30B, several methods cluster within a narrow ECE range, so the result is best read jointly with AUROC. 
CoCA-lite reaches ECE~$0.023$ but has AUROC~$0.508$, consistent with a near-constant confidence predictor that appears calibrated only in aggregate. \textsc{CALIBER} achieves the lowest ECE ($0.022$) while retaining AUROC~$0.797$, separating it from low-ECE but weakly discriminative behavior. DCPO-lite and RLCR obtain marginally higher AUROC ($0.811$ and $0.807$) at the cost of higher ECE. 

Figure~\ref{fig:BigMath_pareto_plot} visualizes this tradeoff. At 7B, \textsc{CALIBER} occupies the most favorable region, combining the lowest ECE with the highest AUROC. At 30B, it shifts furthest toward low ECE while staying competitive in AUROC, separating it from CoCA-lite, whose low ECE is practically uninformative due to poor AUROC.

\begin{table}[h!]
\centering
\small
\begin{tabular}{lcccccccc}
\toprule
& \multicolumn{4}{c}{7B} & \multicolumn{4}{c}{30B} \\
\cmidrule(lr){2-5} \cmidrule(lr){6-9}
Method
& ECE $\downarrow$ & Brier $\downarrow$ & AUROC $\uparrow$ & Acc. $\uparrow$
& ECE $\downarrow$ & Brier $\downarrow$ & AUROC $\uparrow$ & Acc. $\uparrow$ \\
\midrule
Base & 0.208 & 0.208 & 0.644 & 0.761 & 0.137 & 0.144 & 0.609 & 0.751 \\
RLVR & 0.211 & 0.210 & 0.502 & 0.789 &  0.162 & 0.166 &	0.528 &	\textbf{0.833}  \\
RLCR & 0.092 & 0.142 & 0.753 & 0.786 &  0.029 & \textbf{0.110} &  0.807 & 0.826    \\
COREA-lite & 0.079 & 0.142 & 0.793 & 0.793 & 0.039 & 0.114 & 0.787 & 0.827  \\
CoCA-lite & 0.070 & 0.151 & 0.681 & \textbf{0.801} &  0.023 &	0.143	& 0.508 &	0.828 \\
DCPO-lite & 0.059 & 0.136 & 0.780 & 0.789 & 0.027 & 0.115 & \textbf{0.811} & 0.817 \\
\midrule
\textbf{CALIBER} & \textbf{0.028} & \textbf{0.130} & \textbf{0.806} & 0.780 & \textbf{0.022}	& 0.115	& 0.797	& 0.817  \\
\bottomrule
\end{tabular}
\caption{BigMathDigits calibration and accuracy metrics for the 7B and 30B models. \textsc{CALIBER} achieves the lowest ECE at both scales, with the largest gains on the 7B model, while maintaining competitive task accuracy. Best values in each column are shown in \textbf{bold}.}
\label{tab:bigmathdigits-selected-checkpoints}
\end{table}

These calibration gains do not come from substantially sacrificing task accuracy. Relative to standard RLVR, \textsc{CALIBER}'s accuracy changes by $-0.9$ and $-1.6$ percentage points for the 7B and 30B models. Eliciting confidence both before and after thinking thus improves absolute calibration while preserving competitive task performance.

\begin{figure}[t]
    \centering
    \begin{subfigure}[t]{0.49\linewidth}
        \centering
        \includegraphics[width=\linewidth]{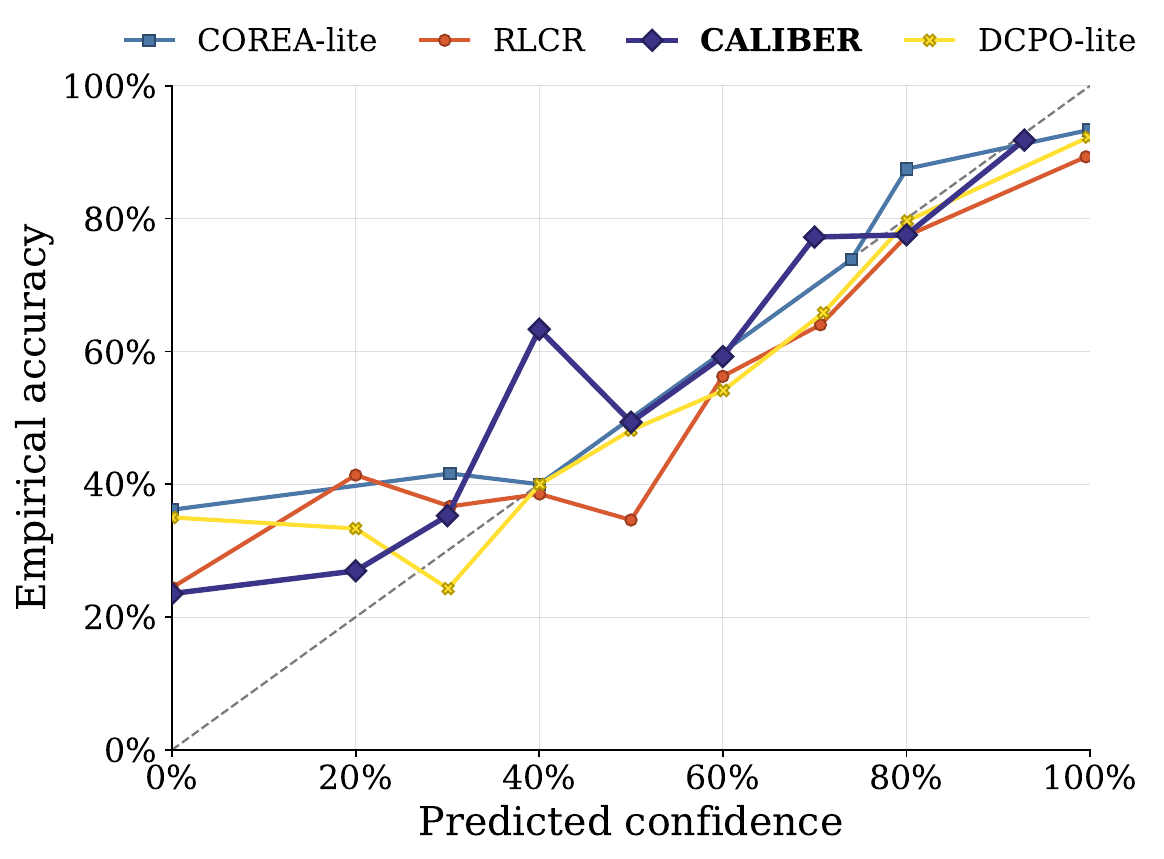}
        \caption{7B model.}
        \label{fig:bigmathdigits_reliability_7b}
    \end{subfigure}
    \hfill
    \begin{subfigure}[t]{0.49\linewidth}
        \centering
        \includegraphics[width=\linewidth]{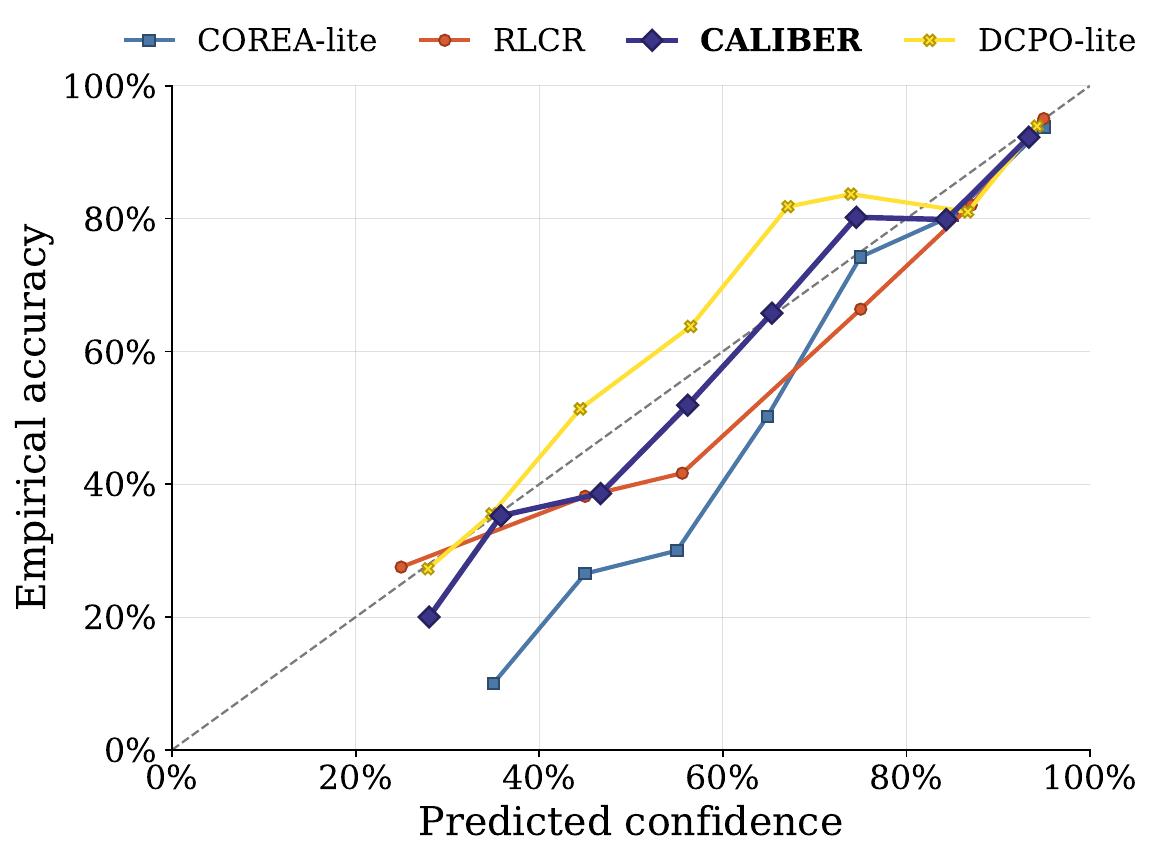}
        \caption{30B model.}
        \label{fig:bigmathdigits_reliability_30b}
    \end{subfigure}
    \caption{Reliability diagrams for BigMathDigits, omitting bins with fewer than 10 samples. Curves closer to the diagonal indicate better absolute calibration. \textsc{CALIBER} avoids severe overconfidence and remains well aligned in high-confidence bins.}
    \label{fig:BigMath_reliability_diagram}
\end{figure}

The reliability diagrams in Figure~\ref{fig:BigMath_reliability_diagram} display the confidence behavior behind the aggregate calibration metrics in more detail. \textsc{CALIBER} generally avoids severe overconfidence and stays well aligned in the high-confidence region, where many selective-prediction decisions are made. Calibration behavior varies across confidence bins, reinforcing why we report ECE alongside Brier score and AUROC rather than relying on any single metric.

Finally, \textsc{CALIBER} uses its reasoning trace and answer to adjust its post-confidence relative to its pre-confidence. For both models, updates are more negative for incorrect answers than for correct answers, indicating that the model meaningfully lowers confidence when the realized answer is incorrect. Thus, the final estimate responds to the realized solution by updating the initial estimate rather than copying it. Per-dataset update values for correct and incorrect responses for both models are reported in Appendix~\ref{app:confidence_updates}.

\subsection{Out-of-Distribution Generalization}

\begin{figure}[t]
    \centering
    \begin{subfigure}[t]{0.49\linewidth}
        \centering
        \includegraphics[width=\linewidth]{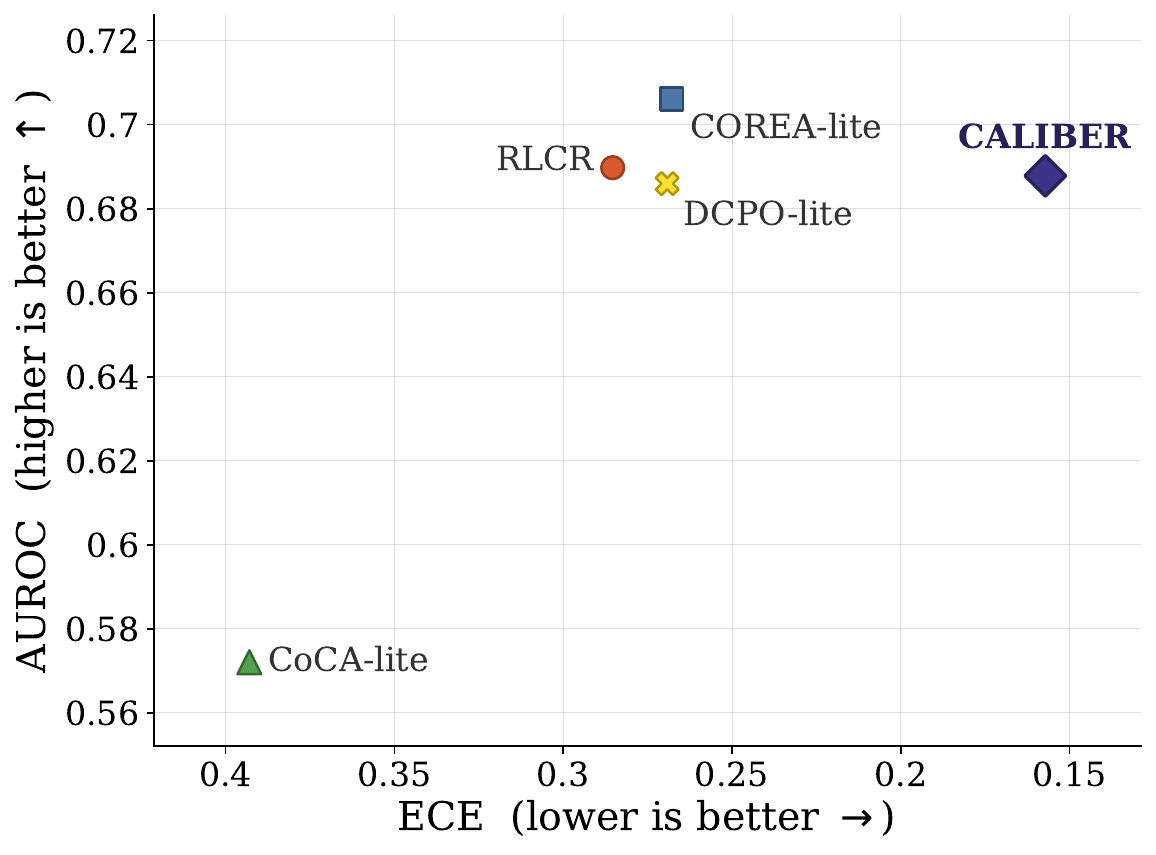}
        \caption{7B model.}
        \label{fig:ood_pareto_7b}
    \end{subfigure}
    \hfill
    \begin{subfigure}[t]{0.49\linewidth}
        \centering
        \includegraphics[width=\linewidth]{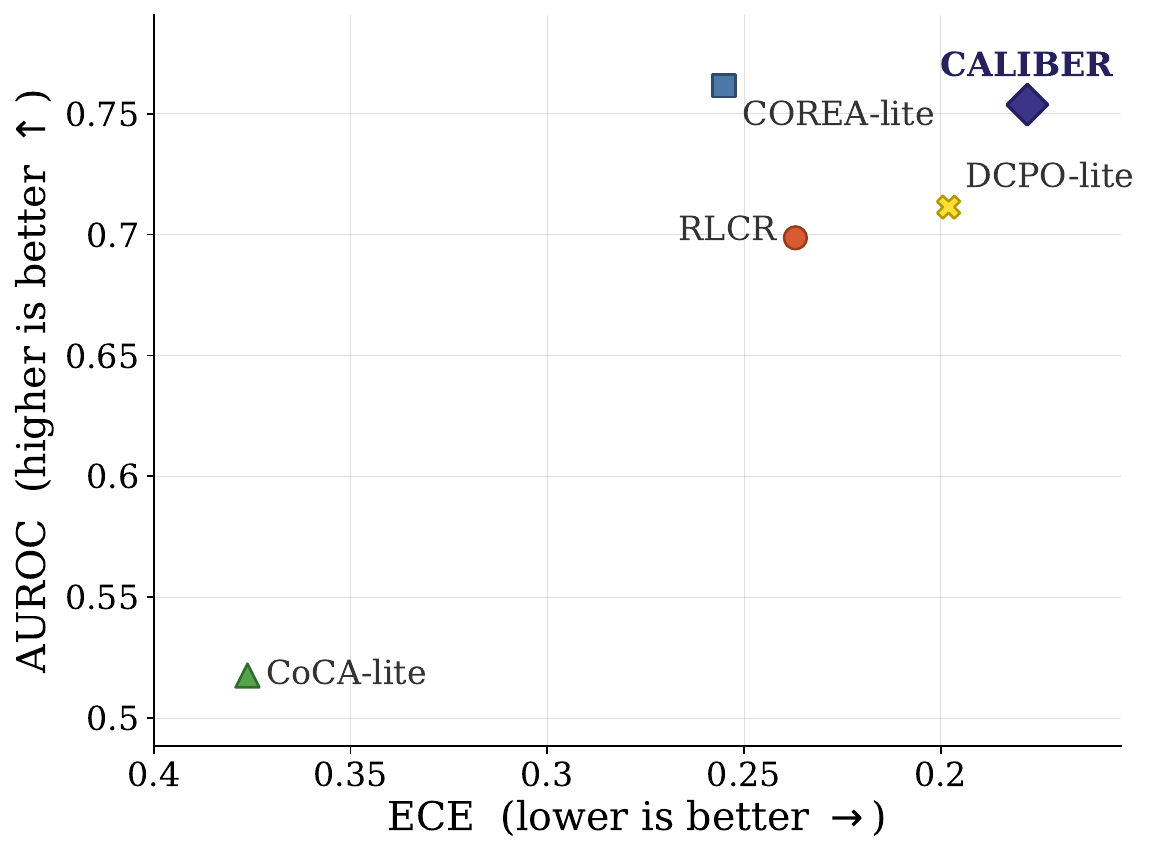}
        \caption{30B model.}
        \label{fig:ood_pareto_30b}
    \end{subfigure}
    
    \caption{Average OOD ECE--AUROC tradeoff across GPQA, TriviaQA, and SimpleQA. Lower ECE and higher AUROC are better. \textsc{CALIBER} remains near the low-ECE/high-AUROC frontier for both model sizes.}
    \label{fig:ood-pareto}
    \vspace{-3mm}
\end{figure}

Calibration under distribution shift is important because confidence estimates that are reliable in the training distribution may not remain reliable on shifted tasks or domains~\citep{ovadia2019can}. We therefore evaluate whether \textsc{CALIBER}'s calibration gains transfer beyond the BigMathDigits training distribution. Results on GPQA, TriviaQA, and SimpleQA are reported in Table~\ref{tab:ood-results-7B} for the 7B model and Table~\ref{tab:ood-results-30B} for the 30B model, while Figure~\ref{fig:ood-pareto} summarizes the average ECE--AUROC tradeoff across the three OOD datasets.

\begin{table*}[h!]
\centering
\scriptsize
\resizebox{\textwidth}{!}{
\begin{tabular}{lcccccccccccc}
\toprule
& \multicolumn{4}{c}{\textbf{GPQA}} 
& \multicolumn{4}{c}{\textbf{TriviaQA}} 
& \multicolumn{4}{c}{\textbf{SimpleQA}} \\
\cmidrule(lr){2-5}
\cmidrule(lr){6-9}
\cmidrule(lr){10-13}
\textbf{Method} 
& ECE $\downarrow$ & Brier $\downarrow$ & AUROC $\uparrow$ & Acc. $\uparrow$
& ECE $\downarrow$ & Brier $\downarrow$ & AUROC $\uparrow$ & Acc. $\uparrow$
& ECE $\downarrow$ & Brier $\downarrow$ & AUROC $\uparrow$ & Acc. $\uparrow$ \\
\midrule
Base 
& 0.419 & 0.415 & 0.584 & 0.357
& 0.276 & 0.269 & 0.763 & 0.629
& 0.720 & 0.585 & 0.592 & 0.051 \\

RLVR 
& 0.609 & 0.609 & 0.503 & 0.386
& 0.317 & 0.303 & 0.674 & 0.645
& 0.789 & 0.687 & 0.627 & 0.049 \\

RLCR 
& 0.149 & 0.265 & 0.601 & 0.387
& 0.161 & 0.198 & 0.814 & 0.636
& 0.546 & 0.399 & 0.655 & 0.049 \\

COREA-lite 
& 0.218 & 0.278 & \textbf{0.622} & 0.379
& 0.121 & 0.177 & \textbf{0.840} & \textbf{0.652}
& 0.464 & 0.311 & 0.657 & 0.053 \\

CoCA-lite 
& 0.387 & 0.384 & 0.553 & \textbf{0.388}
& 0.130 & 0.235 & 0.606 & 0.647
& 0.661 & 0.503 & 0.557 & \textbf{0.054} \\

DCPO-lite & 0.127 & 0.252 & 0.599 & 0.373
& 0.141 & 0.196 & 0.789 & 0.637 & 0.540 & 0.382 & \textbf{0.669} & 0.050\\
\midrule
\textsc{CALIBER} 
& \textbf{0.093} & \textbf{0.246} & 0.585 & 0.382
& \textbf{0.082} & \textbf{0.173} & 0.809 & 0.630
& \textbf{0.296} & \textbf{0.163} & \textbf{0.669} & 0.050 \\
\bottomrule
\end{tabular}
}
\caption{Out-of-distribution calibration and accuracy metrics for the 7B model on GPQA, TriviaQA, and SimpleQA.}
\label{tab:ood-results-7B}
\end{table*}

\begin{table*}[t]
\centering
\scriptsize
\resizebox{\textwidth}{!}{
\begin{tabular}{lcccccccccccc}
\toprule
& \multicolumn{4}{c}{\textbf{GPQA}} 
& \multicolumn{4}{c}{\textbf{TriviaQA}} 
& \multicolumn{4}{c}{\textbf{SimpleQA}} \\
\cmidrule(lr){2-5}
\cmidrule(lr){6-9}
\cmidrule(lr){10-13}
\textbf{Method} 
& ECE $\downarrow$ & Brier $\downarrow$ & AUROC $\uparrow$ & Acc. $\uparrow$
& ECE $\downarrow$ & Brier $\downarrow$ & AUROC $\uparrow$ & Acc. $\uparrow$
& ECE $\downarrow$ & Brier $\downarrow$ & AUROC $\uparrow$ & Acc. $\uparrow$ \\
\midrule
Base 
& 0.248& 0.259& 0.626& 0.540& 0.298& 0.300& 0.597& 0.526& 0.887& 0.849& 0.583& 0.036\\

RLVR 
& 0.400 &	0.399 &	0.503 & 0.599 &
0.418 &	0.418	& 0.509	& 0.574 &
0.942 &	0.936 &	0.506 & \textbf{0.051} \\

RLCR 
& 0.130 & 	0.260&	0.618&	0.428
& 0.144	& 0.197 & 0.808 & 0.562 &
0.437 & 0.268 &	0.670 & 0.049 \\

COREA-lite 
& 0.090&	0.212&	\textbf{0.733}&	0.588
& 0.183	& 0.197	& \textbf{0.868}	& 0.570
& 0.492	 & 0.295  &	\textbf{0.683}  &	0.049\\

CoCA-lite 
& 0.190	& 0.273	& 0.571	& \textbf{0.614}
& 0.183	& 0.269 &	0.495 &	\textbf{0.621}
& 0.756	& 0.618	& 0.487	& 0.049 \\

DCPO-lite 
& 0.100 & 0.232 & 0.697 & 0.557
& 0.119 & 0.177 & 0.839 & 0.549
& \textbf{0.374}& \textbf{0.193} & 0.599 & 0.048 \\

\midrule

\textsc{CALIBER} 
& \textbf{0.025} & \textbf{0.205} & 0.731 & 0.580
& \textbf{0.111}	& \textbf{0.170}	& 0.857	& 0.570
& 0.399	& 0.214	& 0.673	& 0.049 \\

\bottomrule
\end{tabular}
}
\caption{Out-of-distribution calibration and accuracy metrics for the 30B model on GPQA, TriviaQA, and SimpleQA.}
\label{tab:ood-results-30B}
\end{table*}

Across GPQA and TriviaQA, \textsc{CALIBER} consistently improves absolute calibration. For both model sizes, it achieves the lowest ECE and Brier score on these datasets. The AUROC results show a tradeoff similar to the in-domain setting: COREA-lite or DCPO-lite sometimes obtain slightly stronger failure prediction, but at the cost of higher calibration error. Thus, on these OOD datasets, \textsc{CALIBER} provides the most reliable probability estimates while preserving competitive discrimination between correct and incorrect answers.

SimpleQA is a more extreme distribution shift and should be interpreted separately. All methods achieve very low accuracy on this dataset, around five percent, so calibration and discrimination metrics are estimated from relatively few correct responses. For the 7B model, \textsc{CALIBER} achieves the best ECE and Brier score and ties DCPO-lite for the best AUROC. For the 30B model, DCPO-lite obtains lower ECE and Brier score, while \textsc{CALIBER} achieves substantially higher AUROC than DCPO-lite. This suggests that, in the lowest-accuracy regime, methods can occupy different parts of the calibration--discrimination tradeoff.

Figure~\ref{fig:ood-pareto} aggregates this behavior across OOD datasets. \textsc{CALIBER} remains near the favorable low-ECE/high-AUROC region for both model sizes: it improves absolute calibration relative to most baselines while avoiding the weak failure prediction observed for some low-ECE methods. This supports the main OOD conclusion that position--target alignment improves calibration under distribution shift without collapsing confidence into an uninformative average. The full set of Pareto plots for each dataset is available in Appendix~\ref{app:pareto}.

Quantitatively, \textsc{CALIBER} reduces ECE relative to the strongest single-confidence baseline on GPQA and TriviaQA for both model sizes. For the 7B model, ECE drops from $0.127$ to 0.093 on GPQA, from 0.121 to 0.082 on TriviaQA, and from 0.464 to 0.296 on SimpleQA, corresponding to relative reductions of 26.8\%, 32.2\%, and 36.2\%. For the 30B model, ECE drops from 0.090 to 0.025 on GPQA and from 0.119 to 0.111 on TriviaQA. On SimpleQA, DCPO-lite obtains lower ECE than \textsc{CALIBER}, but \textsc{CALIBER} achieves higher AUROC, reflecting a calibration--discrimination tradeoff.

\begin{figure}[t]
    \centering
    \begin{subfigure}[t]{0.49\linewidth}
        \centering
        \includegraphics[width=\linewidth]{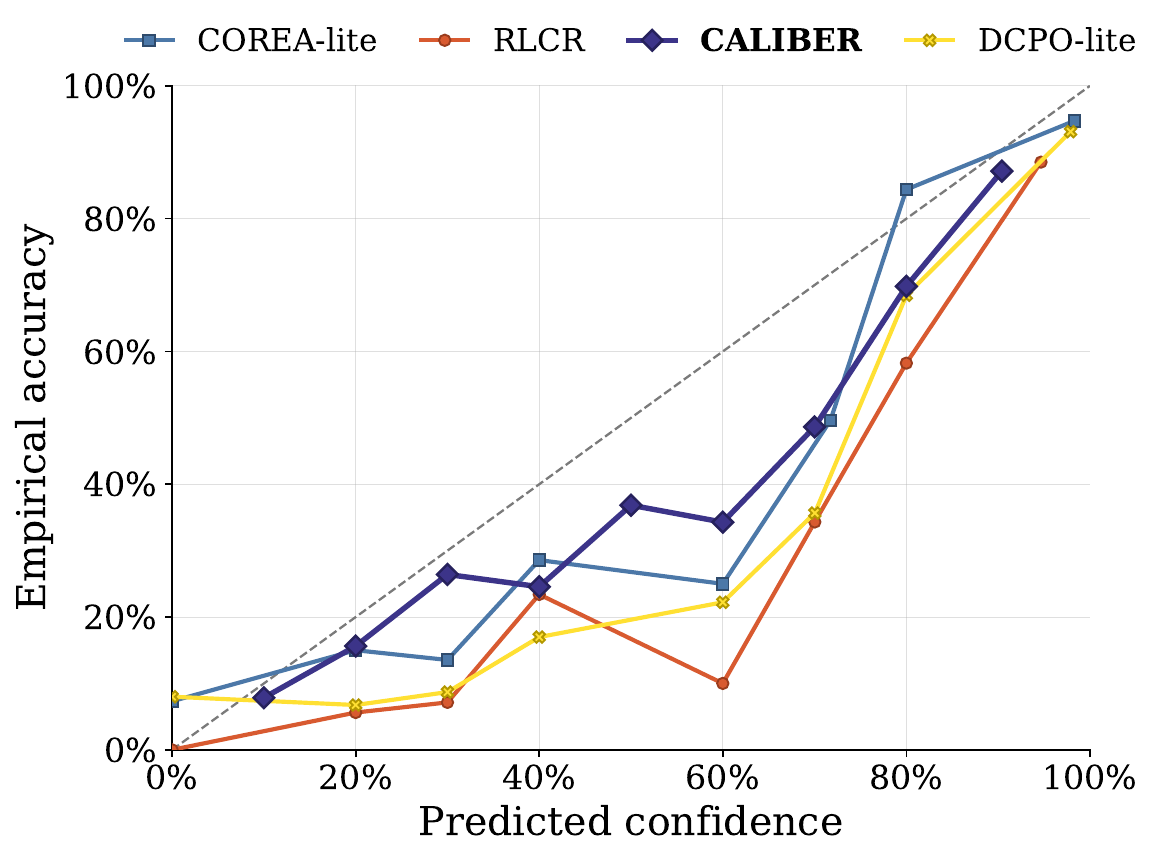}
        \caption{7B model.}
        \label{fig:reliability_triviaqa_7b}
    \end{subfigure}
    \hfill
    \begin{subfigure}[t]{0.49\linewidth}
        \centering
        \includegraphics[width=\linewidth]{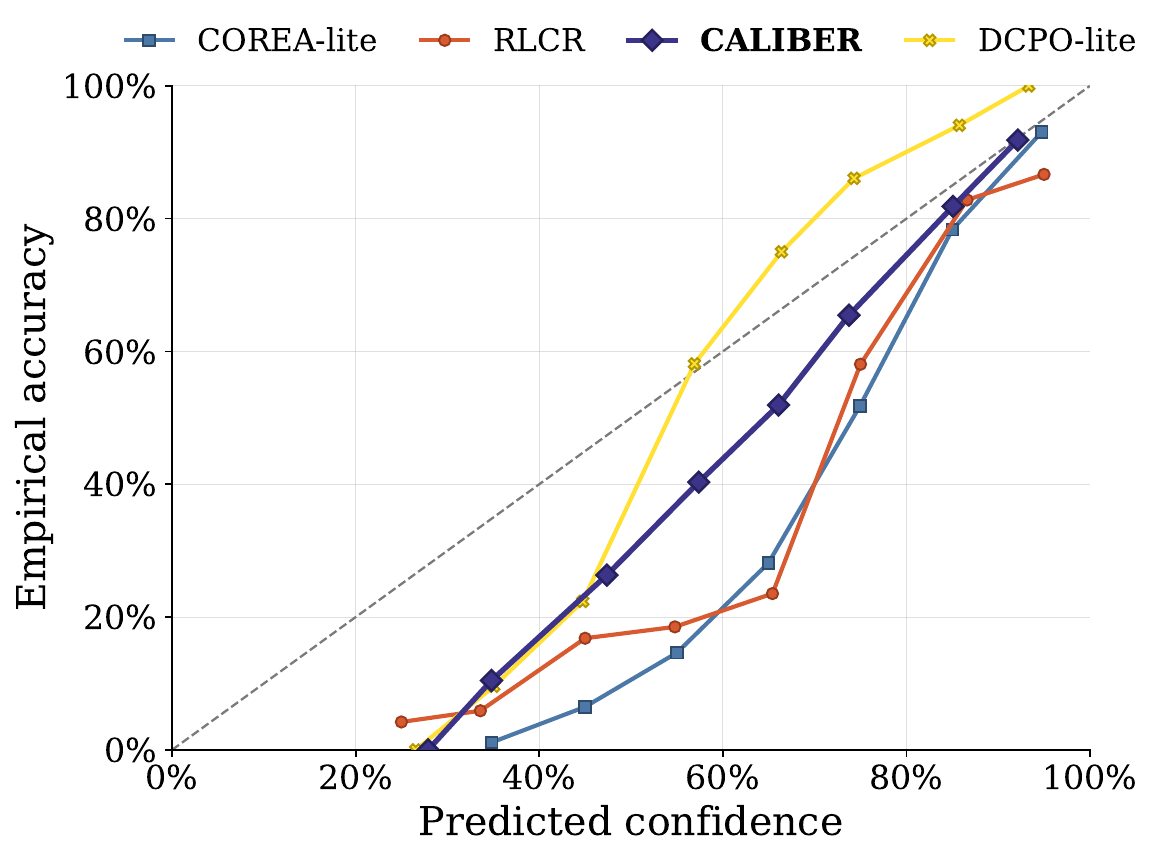}
        \caption{30B model.}
        \label{fig:reliability_triviaqa_30b}
    \end{subfigure}

    \caption{Reliability diagrams on TriviaQA, omitting bins with fewer than 10 samples. Curves closer to the diagonal indicate better absolute calibration. \textsc{CALIBER} generally follows the diagonal more closely than the single-confidence baselines, consistent with its lower ECE and Brier score.}
    \label{fig:reliability_ood_triviaqa}
\end{figure}

The TriviaQA reliability diagrams in Figure~\ref{fig:reliability_ood_triviaqa} provide a complementary view of the aggregate calibration metrics. \textsc{CALIBER}'s curves follow the diagonal more closely across much of the confidence range than the single-confidence baselines, matching its lower ECE and Brier score in Table~\ref{tab:ood-results-7B}. 
The remaining deviations from the diagonal also show that OOD calibration remains challenging, even when the matched pre/post confidence objective improves calibration. The full set of reliability diagrams is available in Appendix~\ref{app:reliability}.

Finally, the confidence-update behavior observed in-domain is preserved out of distribution. On the 7B model, \textsc{CALIBER}'s post-confidence rises by \textbf{4.97 percentage points} on average for correct responses and falls by \textbf{3.61 percentage points} for incorrect responses. Appendix~\ref{app:confidence_updates} reports the per-dataset breakdown and illustrative qualitative examples.

\subsection{Ablation: Position--target Alignment}

\begin{table}[h!]
\centering
\footnotesize
\setlength{\tabcolsep}{3pt}
\begin{tabular}{llcccccccc}
\toprule
& & \multicolumn{4}{c}{7B} & \multicolumn{4}{c}{30B} \\
\cmidrule(lr){3-6} \cmidrule(lr){7-10}
Data & Method
& ECE $\downarrow$ & Brier $\downarrow$ & AUROC $\uparrow$ & Acc. $\uparrow$
& ECE $\downarrow$ & Brier $\downarrow$ & AUROC $\uparrow$ & Acc. $\uparrow$ \\
\midrule
\multirow{2}{*}{BigMathDigits}
& Swapped & \textbf{0.024} & \textbf{0.126} & 0.790 & \textbf{0.793} &  0.087 & 0.127 & 0.739 & \textbf{0.823}  \\
& \textsc{CALIBER} & 0.028 & 0.130 & \textbf{0.806} & 0.780 &  \textbf{0.022}	& \textbf{0.115} &	\textbf{0.797} &	0.817 \\
\midrule
\multirow{2}{*}{GPQA}
& Swapped & 0.123 & \textbf{0.241} & \textbf{0.600} & 0.354 & 0.143	 & 0.227  &	0.725  & \textbf{0.581} \\
& \textsc{CALIBER} & \textbf{0.093} & 0.246 & 0.585 & \textbf{0.382} &  \textbf{0.025} & \textbf{0.205}	& \textbf{0.731} & 0.580 \\
\midrule
\multirow{2}{*}{TriviaQA}
& Swapped & 0.128 & 0.193 & 0.800 & \textbf{0.650} & 0.256	 &  0.235 & 0.840	 & 0.569 \\
& \textsc{CALIBER} & \textbf{0.082} & \textbf{0.173} & \textbf{0.809} & 0.630 &  \textbf{0.111} & \textbf{0.170} & \textbf{0.857} & \textbf{0.570 } \\
\midrule
\multirow{2}{*}{SimpleQA}
& Swapped & 0.529 & 0.354 & 0.640 & 0.046 & 0.536 &	0.356 & \textbf{0.698} & \textbf{0.050}\\
& \textsc{CALIBER} & \textbf{0.296} & \textbf{0.163} & \textbf{0.669} & \textbf{0.050} & \textbf{0.399} & \textbf{0.214} & 0.673 & 0.049\\
\bottomrule
\end{tabular}
\caption{Ablation comparing \textsc{CALIBER} with a \textit{Swapped} objective that reverses the calibration targets for pre- and post-confidence. Best values within each dataset/model pair are shown in \textbf{bold}.}
\label{tab:ablation-ours-swapped}
\vspace{-2mm}
\end{table}

To evaluate the role of position--target alignment, we run an ablation in which the confidence targets are swapped. In the swapped-rewards run, post-confidence is trained with the group-wise target and pre-confidence is trained with the instance-wise target. 
To mirror CALIBER's schedule, we first train with the COREA-lite objective for 400 steps, which corresponds to applying a warmup to the component carrying the group-wise target, before training for a further 100 steps with the full objective. 
Table~\ref{tab:ablation-ours-swapped} compares this swapped configuration against \textsc{CALIBER} across all evaluation datasets.

\textbf{In-domain, the effect of alignment depends on model scale.}
For the 30B model, \textsc{CALIBER} substantially improves over Swapped, reducing ECE from 0.087 to 0.022, reducing Brier score from 0.127 to 0.115, and increasing AUROC from 0.739 to 0.797. For the 7B model, the result is more mixed: Swapped achieves slightly lower ECE and Brier score, while \textsc{CALIBER} achieves higher AUROC. This suggests that in-domain performance alone does not fully distinguish the two objectives.

\textbf{The clearest advantage of position--target alignment appears under distribution shift.}
Across all three OOD datasets and both model sizes, \textsc{CALIBER} achieves lower ECE than Swapped. The largest gaps appear for the 30B model, where alignment reduces ECE by 0.118 on GPQA, 0.145 on TriviaQA, and 0.137 on SimpleQA. The gains also often extend beyond ECE: \textsc{CALIBER} improves Brier score on all OOD datasets except GPQA for the 7B model, and improves AUROC on all OOD datasets except GPQA for the 7B model and SimpleQA for the 30B model. Overall, the ablation supports the main hypothesis that matching each confidence estimate to the target appropriate for its information state is especially important for OOD calibration.

\section{Conclusion}
\label{sec:conclusion}

We introduced \textsc{CALIBER}, a state-dependent calibration framework for reasoning language models. 
Rather than treating confidence as a single scalar attached to an answer, \textsc{CALIBER} elicits confidence before and after thinking and supervises each estimate with the target appropriate to its information state. Pre-confidence is trained to estimate prompt-level success under the current policy, while post-confidence is trained to estimate the correctness of the realized answer. 
Across in-domain and out-of-distribution evaluations, \textsc{CALIBER} improves calibration while maintaining competitive task performance, The swapped-target ablation further supports the central claim that matching confidence position to supervision target is especially important under distribution shift.

More broadly, our results suggest that calibration for reasoning models should be treated as a dynamic problem. One possible explanation for \textsc{CALIBER}'s stronger out-of-distribution calibration is that the pre-confidence objective encourages the model to form an explicit prompt-level estimate of solvability before committing to a reasoning path. Because this estimate is conditioned only on the prompt and supervised with a group-wise success target, it may capture input-level cues associated with domain, format, or difficulty, without conflating them with evidence from a particular generated answer. The post-confidence estimate can then specialize in assessing the realized reasoning trace and answer. The gap between pre- and post-confidence therefore provides a simple diagnostic of how uncertainty changes during reasoning, and opens the door to richer forms of confidence elicitation in long reasoning traces, tool use, and multi-turn interactions. Future work should extend \textsc{CALIBER} beyond this two-point setting, test this prompt-level solvability hypothesis more directly, combine \textsc{CALIBER} with method-specific enhancements such as segmented credit assignment or adaptive reward weighting, and validate state-dependent confidence across additional model families, training distributions, and task domains.



\bibliography{paper,anthology}

\clearpage
\appendix

\section{Appendix}
\label{sec:appendix}

\subsection{Training and Evaluation Details}
\label{app:training_details}

Table~\ref{tab:hyperparams} summarizes the main training and evaluation settings used for the 7B and 30B experiments.

\begin{table}[h!]
\centering
\small
\begin{tabular}{lc}
\toprule
\textbf{Hyperparameter / setting} & \textbf{Value} \\
\midrule
\multicolumn{2}{l}{\textit{Shared training settings}} \\
Calibration RL steps                 & 500 \\
Reward weights                       & Equal \\
Training batch size                  & 2048 \\
Rollouts per prompt during training  & 32 \\
Max sequence length                &  16384\\

\midrule
\multicolumn{2}{l}{\textit{Model-specific training settings (7B / 30B)}} \\
\textsc{Caliber} warmup steps        & 400 / 100 \\
Second-stage joint-training steps    & 100 / 400 \\
Learning rate                        & 7.5e-7 / 4.0e-6 \\
\midrule
\multicolumn{2}{l}{\textit{Shared evaluation settings}} \\
Sampling temperature     & 0.3 \\
Top-$p$                  & 0.75 \\
Malformed-generation filtering      & Yes \\
Repetitive-loop filtering           & Yes \\
Responses per question, non-GPQA     & 3 \\
Responses per question, GPQA         & 12 \\
ECE bins                             & 10 \\
\bottomrule
\end{tabular}
\caption{Training and evaluation hyperparameters. Model-specific values are given as 7B / 30B.}
\label{tab:hyperparams}
\end{table}

\subsection{Per-Dataset Comparison Against the Strongest Baseline}
\label{app:best-baseline}

Tables~\ref{tab:summary-ece} and~\ref{tab:summary-auroc} compare \textsc{CALIBER} to the strongest single-confidence baseline on each dataset, separately for ECE and AUROC and for both model sizes. The strongest baseline is selected per metric, model, and dataset, so the named baseline can differ across cells. On ECE,  \textsc{CALIBER} is best on all four benchmarks for 7B and three of four for 30B. On AUROC,  \textsc{CALIBER} is best or tied on two of four benchmarks for 7B, and for 30B remains within $0.014$ of the strongest baseline on every dataset. No baseline beats \textsc{CALIBER} across both AUROC and ECE. This is consistent with the joint-metric framing in our paper: ECE and AUROC measure different properties, and methods that improve only one can degrade the other. Full results for all methods are in Tables~\ref{tab:bigmathdigits-selected-checkpoints}, ~\ref{tab:ood-results-7B}, and ~\ref{tab:ood-results-30B}.

\begin{table}[h!]
\centering
\small
\setlength{\tabcolsep}{4pt}
\resizebox{\textwidth}{!}{
\begin{tabular}{lcccccccc}
\toprule
& \multicolumn{4}{c}{7B} & \multicolumn{4}{c}{30B} \\
\cmidrule(lr){2-5} \cmidrule(lr){6-9}
Dataset & Baseline & \textsc{CALIBER} & $\Delta$ & Rel. & Baseline & \textsc{CALIBER} & $\Delta$ & Rel. \\
\midrule
BigMathDigits & 0.059 & \textbf{0.028} & $-0.031$ & $-52.5\%$ & 0.023 & \textbf{0.022} & $-0.001$ & $-4.3\%$ \\
GPQA          & 0.127 & \textbf{0.093} & $-0.034$ & $-26.8\%$ & 0.090 & \textbf{0.025} & $-0.065$ & $-72.2\%$ \\
TriviaQA      & 0.121 & \textbf{0.082} & $-0.039$ & $-32.2\%$ & 0.119 & \textbf{0.111} & $-0.008$ & $-6.7\%$ \\
SimpleQA      & 0.464 & \textbf{0.296} & $-0.168$ & $-36.2\%$ & \textbf{0.374} & 0.399 & $+0.025$ & $+6.7\%$ \\
\bottomrule
\end{tabular}
}
\caption{ECE ($\downarrow$) comparison between \textsc{CALIBER} and the strongest single-confidence baseline per dataset, for the 7B and 30B models. 7B baselines: DCPO-lite (BigMathDigits, GPQA), COREA-lite (TriviaQA, SimpleQA). 30B baselines: CoCA-lite (BigMathDigits), COREA-lite (GPQA), DCPO-lite (TriviaQA, SimpleQA). Best per cell in bold.}
\label{tab:summary-ece}
\end{table}
\begin{table}[h!]
\centering
\small
\setlength{\tabcolsep}{4pt}
\resizebox{\textwidth}{!}{
\begin{tabular}{lcccccccc}
\toprule
& \multicolumn{4}{c}{7B} & \multicolumn{4}{c}{30B} \\
\cmidrule(lr){2-5} \cmidrule(lr){6-9}
Dataset & Baseline & \textsc{CALIBER} & $\Delta$ & Rel. & Baseline & \textsc{CALIBER} & $\Delta$ & Rel. \\
\midrule
BigMathDigits & 0.793 & \textbf{0.806} & $+0.013$ & $+1.6\%$ & \textbf{0.811} & 0.797 & $-0.014$ & $-1.7\%$ \\
GPQA          & \textbf{0.622} & 0.585 & $-0.037$ & $-5.9\%$ & \textbf{0.733} & 0.731 & $-0.002$ & $-0.3\%$ \\
TriviaQA      & \textbf{0.840} & 0.809 & $-0.031$ & $-3.7\%$ & \textbf{0.868} & 0.857 & $-0.011$ & $-1.3\%$ \\
SimpleQA      & \textbf{0.669} & \textbf{0.669} & $+0.000$ & $+0.0\%$ & \textbf{0.683} & 0.673 & $-0.010$ & $-1.5\%$ \\
\bottomrule
\end{tabular}
}
\caption{AUROC ($\uparrow$) comparison between \textsc{CALIBER} and the strongest single-confidence baseline per dataset, for the 7B and 30B models. 7B baselines: COREA-lite (BigMathDigits, GPQA, TriviaQA), DCPO-lite (SimpleQA). 30B baselines: DCPO-lite (BigMathDigits), COREA-lite (GPQA, TriviaQA, SimpleQA). Best per cell in bold.}
\label{tab:summary-auroc}
\end{table}

\subsection{Confidence Updates}
\label{app:confidence_updates}

Training with \textsc{CALIBER} results in a model that increases its confidence more after reasoning on questions it ends up getting correct than on those it answers incorrectly. This demonstrates that the model is using the additional context of its thinking trace and answer effectively to update its confidence.

This is shown in more detail in Table \ref{tab:confidence-update-by-dataset}, which lists the \textsc{CALIBER}-trained model's average pre- and post-thinking confidence for each dataset, separated based on correctness. In all cases, the change in confidence after thinking is substantially more positive for correct answers than for incorrect ones.

\begin{table}[h!]
\centering
\footnotesize
\setlength{\tabcolsep}{4pt}
\begin{tabular}{llccc ccc}
\toprule
& & \multicolumn{3}{c}{7B Model} & \multicolumn{3}{c}{30B Model} \\
\cmidrule(lr){3-5} \cmidrule(lr){6-8}
Data & Subset & Pre & Post & $\Delta$ & Pre & Post & $\Delta$ \\
\midrule
\multirow{2}{*}{BigMathDigits}
& Correct   & 0.785 & 0.835 & $+0.050$ & 0.936 & 0.868 & $-0.068$ \\
& Incorrect & 0.626 & 0.545 & $-0.081$ & 0.912 & 0.646 & $-0.266$ \\
\midrule
\multirow{2}{*}{GPQA}
& Correct   & 0.528 & 0.466 & $-0.062$ & 0.890 & 0.648 & $-0.242$ \\
& Incorrect & 0.510 & 0.408 & $-0.102$ & 0.888 & 0.504 & $-0.384$ \\
\midrule
\multirow{2}{*}{TriviaQA}
& Correct   & 0.641 & 0.813 & $+0.171$ & 0.899 & 0.797 & $-0.102$ \\
& Incorrect & 0.514 & 0.540 & $+0.027$ & 0.885 & 0.525 & $-0.360$ \\
\midrule
\multirow{2}{*}{SimpleQA}
& Correct   & 0.429 & 0.468 & $+0.040$ & 0.878 & 0.534 & $-0.344$ \\
& Incorrect & 0.372 & 0.339 & $-0.033$ & 0.876 & 0.443 & $-0.434$ \\
\bottomrule
\end{tabular}
\caption{Mean pre- and post-thinking confidence for \textsc{CALIBER} on correct and incorrect responses, by dataset and model size.}
\label{tab:confidence-update-by-dataset}
\end{table}

We also list several illustrative examples of the 7B model updating its confidence in light of the realized thinking trace and answer -- one correct and one incorrect response from evaluation dataset. These examples illustrate the qualitative behavior encouraged by CALIBER. In the correct examples, the model increases its confidence after thinking through the problem and, conversely, in the incorrect examples, the model decreases its confidence in light of its thinking trace and answer. More generally, as shown in Table~\ref{tab:confidence-update-by-dataset}, confidence updates are more favorable for correct responses than for incorrect responses, indicating that the final estimate responds to the generated reasoning trace and answer rather than simply copying the initial estimate.

\definecolor{calok}{HTML}{1B5E20}
\definecolor{calbad}{HTML}{B71C1C}
\definecolor{calbg}{HTML}{F7F7F7}
\definecolor{calmuted}{HTML}{4A4A4A}

\newcommand{\confcheck}{\textcolor{calok}{\ensuremath{\checkmark}}}
\newcommand{\confcross}{\textcolor{calbad}{\ensuremath{\boldsymbol{\times}}}}

\newcommand{\confupdate}[3]{%
  \textcolor{black}{%
    \ensuremath{p_{\text{pre}}{=}#1 \,\rightarrow\, p_{\text{post}}{=}#2}%
    \;\;\ensuremath{(\Delta{=}#3)}%
  }%
}

\newtcolorbox{calbox}[1][]{%
  enhanced, breakable,
  colback=calbg, colframe=calbg,
  boxrule=0pt, arc=0pt, sharp corners,
  borderline west={1.6pt}{0pt}{calmuted},
  left=7pt, right=4pt, top=3pt, bottom=3pt,
  before skip=4pt, after skip=4pt,
  fontupper=\small,
  #1
}

\newcommand{\calexampleup}[6]{%
  \begin{calbox}[borderline west={1.6pt}{0pt}{calok}]
    \textbf{Q.}~#1\par
    \textbf{A.}~#2~\confcheck\par
    \confupdate{#3}{#4}{#5}\par
    \smallskip
    {\color{calmuted}\itshape #6}
  \end{calbox}%
}

\newcommand{\calexampledown}[6]{%
  \begin{calbox}[borderline west={1.6pt}{0pt}{calbad}]
    \textbf{Q.}~#1\par
    \textbf{A.}~#2~\confcross\par
    \confupdate{#3}{#4}{#5}\par
    \smallskip
    {\color{calmuted}\itshape #6}
  \end{calbox}%
}

\subsubsection{BigMathDigits}

\calexampleup
  {There are 20 points on a circle, paired and connected by 10 chords
   which do not share endpoints and do not intersect. In how many ways
   can this be done?}
  {16796}{0.30}{0.70}{+0.40}
  {``This seems like a combinatorial problem, maybe related to Catalan
numbers? I remember that Catalan numbers count the number of
non-crossing partitions, like parentheses or triangulations.\
{\textbf{[\ldots]}}\ For $n$ pairs, the $n$-th Catalan number gives the
number of non-crossing chord diagrams. So $n = 10$. Let me check:
$C_{10} = \tfrac{1}{11}\binom{20}{10}$. $\binom{20}{10}$ is 184756, so
184756 divided by 11 is 16796.\ {\textbf{[\ldots]}}\ But wait, I should
make sure that this applies here. The problem states that the chords
do not share endpoints and do not intersect. That's exactly the
condition for Catalan numbers.\ {\textbf{[\ldots]}}\ So $C_{10}$ is
indeed 16796.''}

\calexampledown
  {A person contracted the flu, and after two rounds of infection, a total of
   196 people were infected. Calculate the average number of people infected by
   one person in each round of infection. Express your answer as a whole number.}
  {14}{0.80}{0.30}{-0.50}
  {``After two rounds a total of 196 people are infected, and I need the average
number each person infects per round. Let the first person infect $x$ people in
round one; then each of those infects $x$ more in round two, so the total is
$1 + x + x^2 = 196$.\ {\textbf{[\ldots]}}\ Solving $x^2 + x - 195 = 0$ gives
$x \approx 13.5$, which isn't a whole number.\ {\textbf{[\ldots]}}\ Let me just
test values: $k = 13$ gives $1 + 13 + 169 = 183$, and $k = 14$ gives
$1 + 14 + 196 = 211$ -- neither equals 196.\ {\textbf{[\ldots]}}\ The answer is
supposed to be a whole number, so I'm not certain, but it's likely 14.''}

\subsubsection{TriviaQA}

\calexampleup
  {Della Street was whose secretary?}
  {Perry Mason}{0.20}{0.90}{+0.70}
  {``Okay, let's tackle this question: "Della Street was whose secretary?" Hmm, Della Street... I think she's a character from a TV show. Wait, who was her boss? Oh right, she was the secretary to Perry Mason, the famous defense attorney in the TV series "Perry Mason." Della Street was played by Barbara Hale. So the answer should be Perry Mason. But wait, I need to make sure I'm not mixing up any details. Let me double-check. Yes, Della Street is indeed Perry Mason's secretary. I'm pretty confident now.''}

\calexampledown
  {In 2009, who was convicted of the murder of Lana Clarkson?}
  {David Lee Roth}{0.80}{0.20}{-0.60}
  {``Okay, so I need to figure out who was convicted of the murder of Lana Clarkson in 2009. Let me start by recalling what I know about this case. \textbf{[\ldots]} So the most likely scenario is that the user has the wrong year, but if I have to answer based on the given information, the correct answer would still be David Lee Roth, even though the year is incorrect. However, since the user specified 2009, maybe there's a different person. Alternatively, maybe the conviction was in 2009 for a different case. But I don't have information about that. So I think the best answer is David Lee Roth, even though the year is off.''}

\subsubsection{GPQA}

\calexampleup
  {You tag the inner capsid proteins of rotavirus with GFP and deliver them to
   mice via oral gavage. You identify the fluorescent Peyer patches and isolate
   the proliferating cell population. You sequence their variable heavy chain
   gene and observe high variability. Which process is taking place?\\
   A) class switching recombination\quad B) somatic hypermutation\\
   C) complement activation\quad D) VDJ recombination}
  {B}{0.60}{0.90}{+0.30}
  {``High variability in the variable heavy-chain genes of a proliferating B-cell
population in the Peyer's patches points to the germinal-center reaction. Class
switching changes the constant region, not the variable region, so not A. V(D)J
recombination creates the initial diversity in the bone marrow, not in secondary
lymphoid organs, so not D. Complement activation is unrelated to receptor-gene
variability, so not C. Somatic hypermutation introduces point mutations in the
variable region during affinity maturation in germinal centers -- exactly what is
seen here. So the answer is B.''}

\calexampledown
  {An electron is in the spin state $(3i, 4)$. Find the expectation value of its
   spin along the $y$-direction, $S_y$, where
   $\sigma_y=\left(\begin{smallmatrix}0&-i\\ i&0\end{smallmatrix}\right)$.\\
   A) $-12\hbar/25$\quad B) $12\hbar/25$\quad C) $-25\hbar/2$\quad D) $25\hbar/2$}
  {$-25\hbar/2$ (C)}{0.80}{0.30}{-0.50}
  {``$S_y=(\hbar/2)\sigma_y$ and $|\psi\rangle=[3i,4]^{T}$. Computing
$\sigma_y|\psi\rangle=[-4i,-3]^{T}$, then
$\langle\psi|\sigma_y|\psi\rangle=(3i)(-4i)+4(-3)=12-12=0$, so $\langle S_y\rangle=0$.\
{\textbf{[\ldots]}}\ But none of the options is zero, so something is off.\
{\textbf{[\ldots]}}\ If instead this were $S_z$ I would get $-25\hbar/2$, which is
option C.\ {\textbf{[\ldots]}}\ I'll assume there is a typo and go with
C) $-25\hbar/2$.''}

\subsubsection{SimpleQA}

\calexampleup
  {What type of voice did Bessie Smith have?}
  {contralto}{0.20}{0.80}{+0.60}
  {``Okay, let's tackle this question: "What type of voice did Bessie Smith have?" First, I need to recall who Bessie Smith is. I think she's a famous singer, maybe from the jazz or blues era. Wait, Bessie Smith is often referred to as the "Empress of the Blues." Now, about her voice type. Voice types are typically categorized as soprano, contralto, mezzo-soprano, tenor, baritone, etc. Since she's a blues singer, her voice would likely be a contralto because that's common in blues and early jazz. But I should double-check. Maybe someone else with the same name? No, I'm pretty sure it's the same person. So, contralto is the likely answer. But I'm not 100\% certain. Maybe I should confirm by thinking of other contralto singers from that era, like Ma Rainey or Billie Holiday. Bessie Smith's voice was known for its powerful, gravelly quality, which fits a contralto. Yeah, I think that's right.''}

\calexampledown
  {In which year was the municipality of San Eduardo, Boyacá, Colombia, founded?}
  {1730}{0.80}{0.30}{-0.50}
  {``Okay, the user wants the founding year of San Eduardo, a municipality in
Boyacá, Colombia. I don't have a specific founding date stored for this
place.\ {\textbf{[\ldots]}}\ Many towns in Boyacá have Spanish colonial origins,
but some Colombian municipalities were only created in the 20th century, so the
plausible range is wide.\ {\textbf{[\ldots]}}\ I can't verify this against any
source, so I'll fall back on an educated guess based on typical colonial-era
founding dates --- maybe around 1730 --- but I'm really not confident.''}

\subsection{Reliability Diagrams}
\label{app:reliability}

\begin{figure}[h]
    \centering
    \begin{subfigure}[t]{0.49\linewidth}
        \centering
        \includegraphics[width=\linewidth]{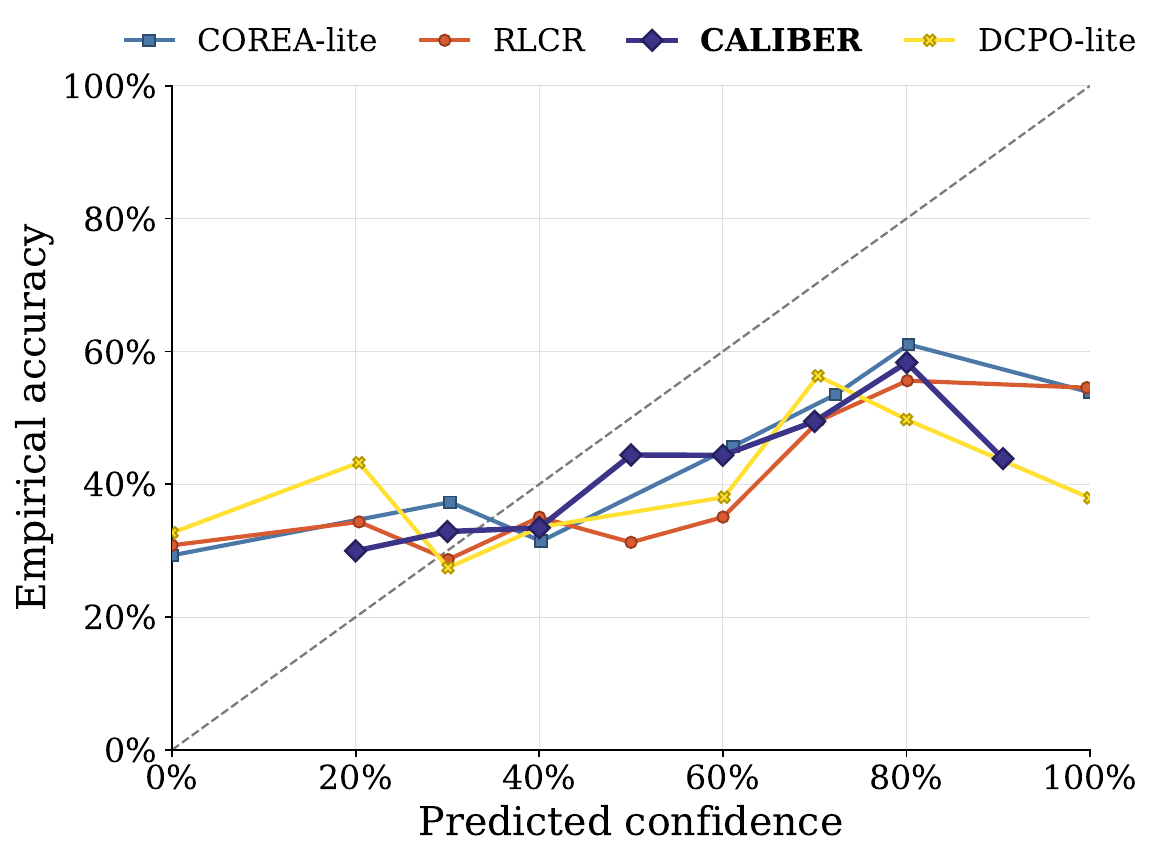}
        \caption{7B model.}
        \label{app:reliability_gpqa_7b}
    \end{subfigure}
    \hfill
    \begin{subfigure}[t]{0.49\linewidth}
        \centering
        \includegraphics[width=\linewidth]{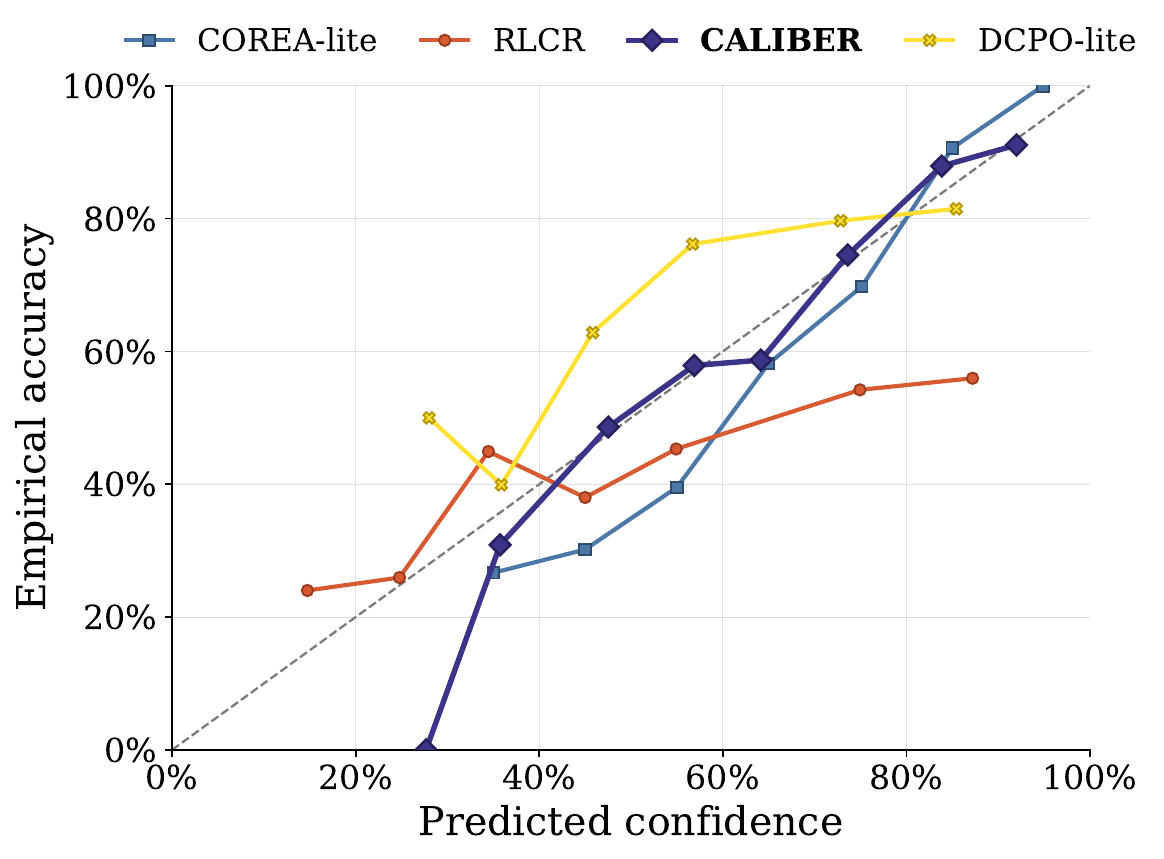}
        \caption{30B model.}
        \label{app:reliability_gpqa_30b}
    \end{subfigure}

    \caption{Reliability diagrams on GPQA.}
    \label{app:reliability_ood_gpqa}
\end{figure}

\begin{figure}[h]
    \centering
    \begin{subfigure}[t]{0.49\linewidth}
        \centering
        \includegraphics[width=\linewidth]{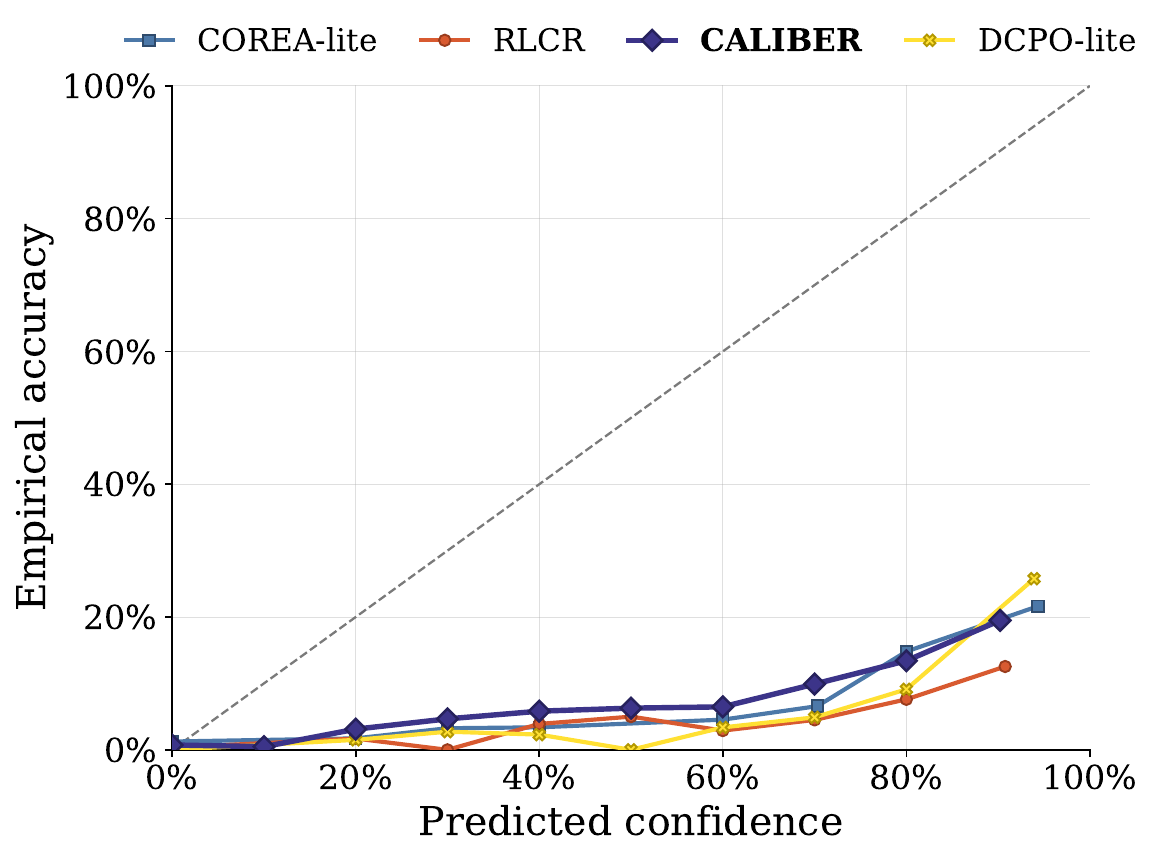}
        \caption{7B model.}
        \label{app:reliability_simpleqa_7b}
    \end{subfigure}
    \hfill
    \begin{subfigure}[t]{0.49\linewidth}
        \centering
        \includegraphics[width=\linewidth]{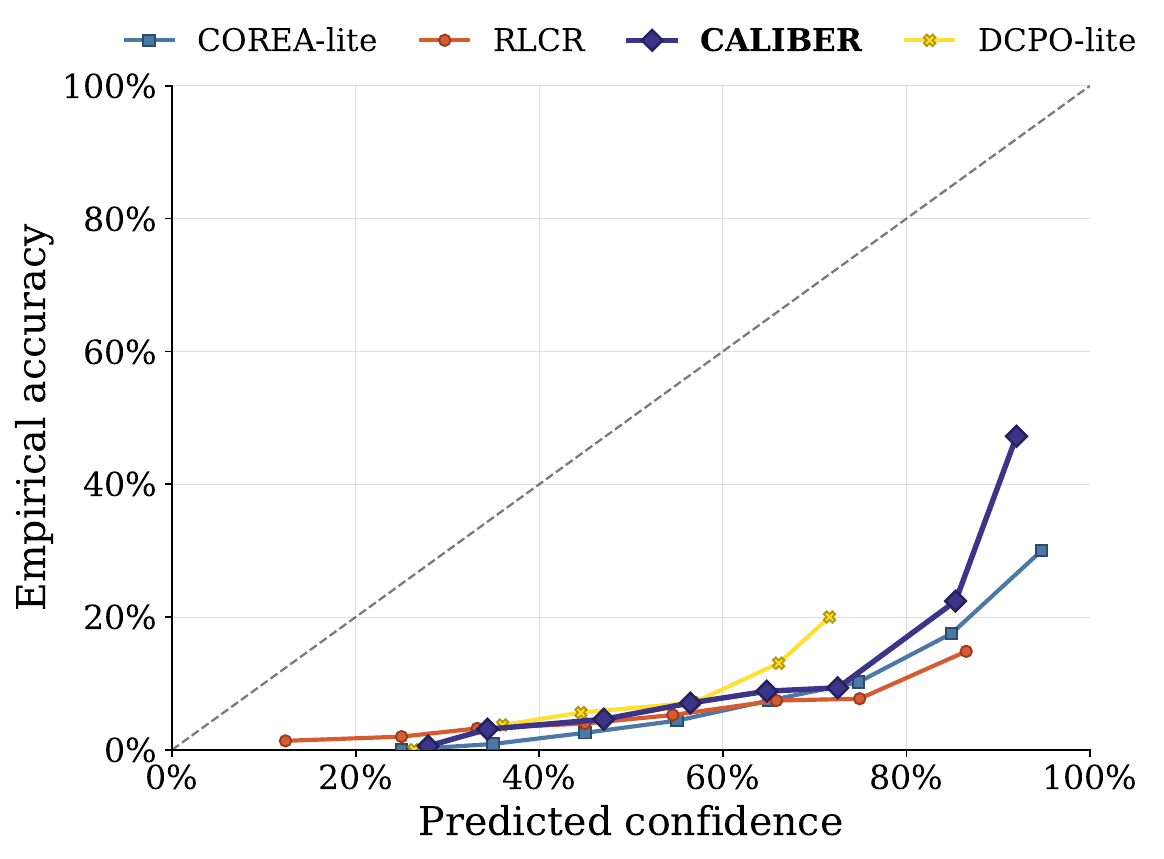}
        \caption{30B model.}
        \label{app:reliability_simpleqa_30b}
    \end{subfigure}

    \caption{Reliability diagrams on SimpleQA.}
    \label{app:reliability_ood_simpleqa}
\end{figure}

Figures~\ref{app:reliability_ood_gpqa} and \ref{app:reliability_ood_simpleqa} show reliability diagrams for \textsc{CALIBER} and the baselines on the remaining evaluation datasets (GPQA and SimpleQA) for both the 7B and 30B models. CoCA-lite is omitted as its confidence estimates all fall into one or two bins, making it impossible to form a true curve. 

Each curve plots empirical accuracy against predicted confidence per bin, so a perfectly calibrated model lies on the diagonal, with points below indicating overconfidence and points above indicating underconfidence. We omit bins with fewer than 10 samples. Across datasets and model sizes, \textsc{CALIBER} generally tracks the diagonal more closely than the single-confidence baselines.

\newpage

\subsection{Pareto Plots}
\label{app:pareto}

\begin{figure}[h]
    \centering
    \begin{subfigure}[t]{0.49\linewidth}
        \centering
        \includegraphics[width=\linewidth]{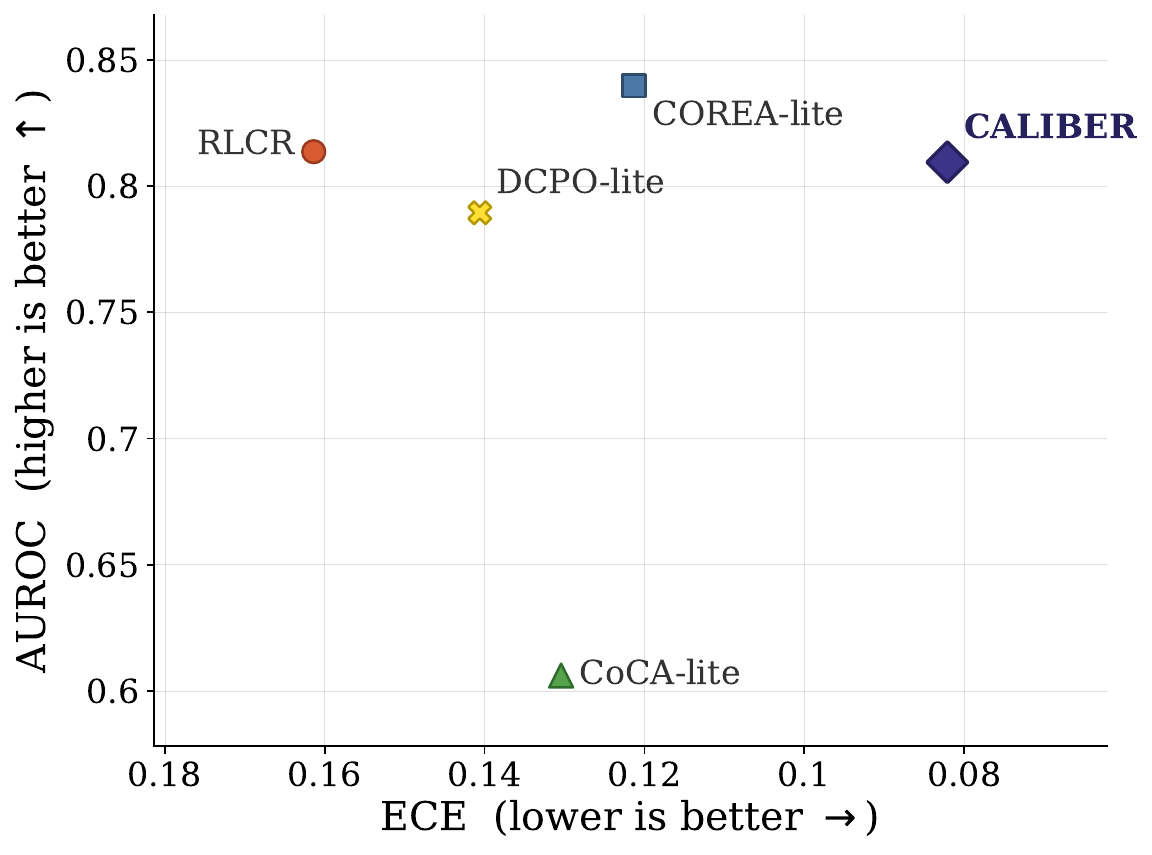}
        \caption{7B model.}
        \label{app:pareto_triviaqa_7b}
    \end{subfigure}
    \hfill
    \begin{subfigure}[t]{0.49\linewidth}
        \centering
        \includegraphics[width=\linewidth]{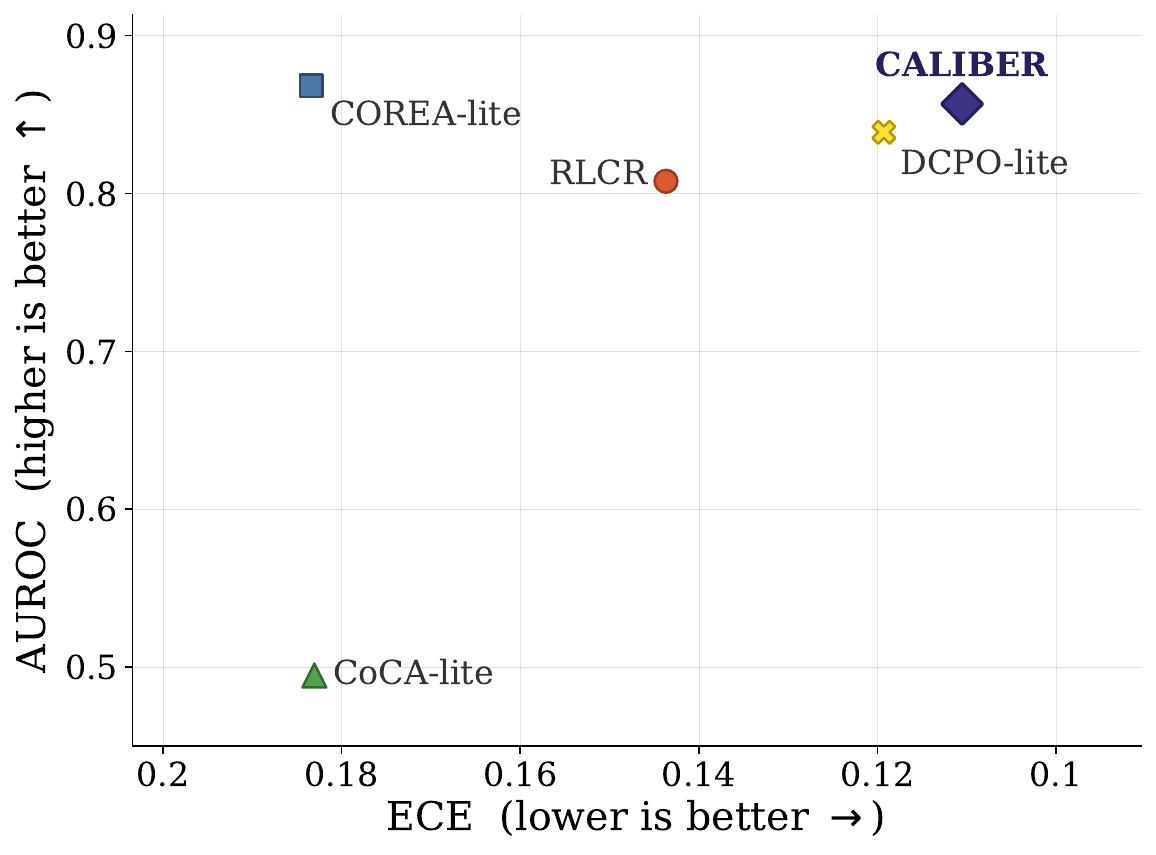}
        \caption{30B model.}
        \label{app:pareto_triviaqa_30b}
    \end{subfigure}

    \caption{Pareto plots on TriviaQA.}
    \label{app:pareto_ood_triviaqa}
\end{figure}

\begin{figure}[h]
    \centering
    \begin{subfigure}[t]{0.49\linewidth}
        \centering
        \includegraphics[width=\linewidth]{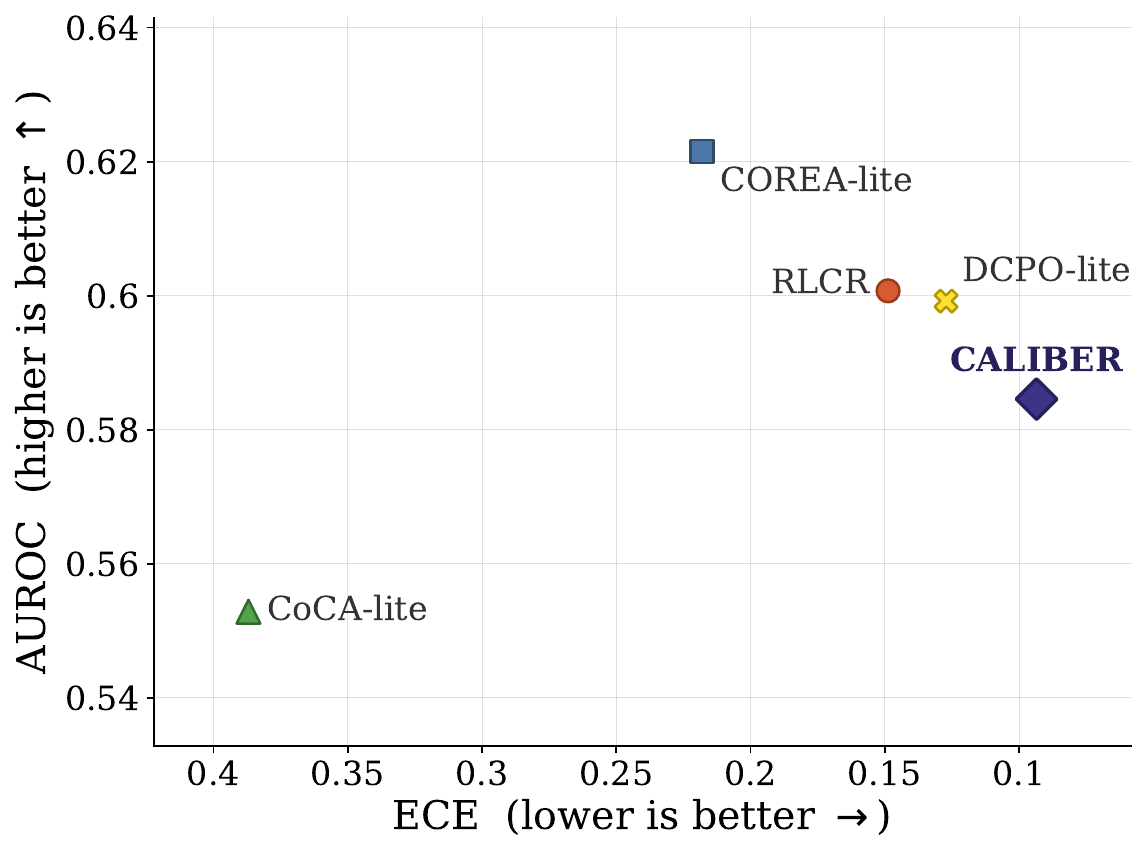}
        \caption{7B model.}
        \label{app:pareto_gpqa_7b}
    \end{subfigure}
    \hfill
    \begin{subfigure}[t]{0.49\linewidth}
        \centering
        \includegraphics[width=\linewidth]{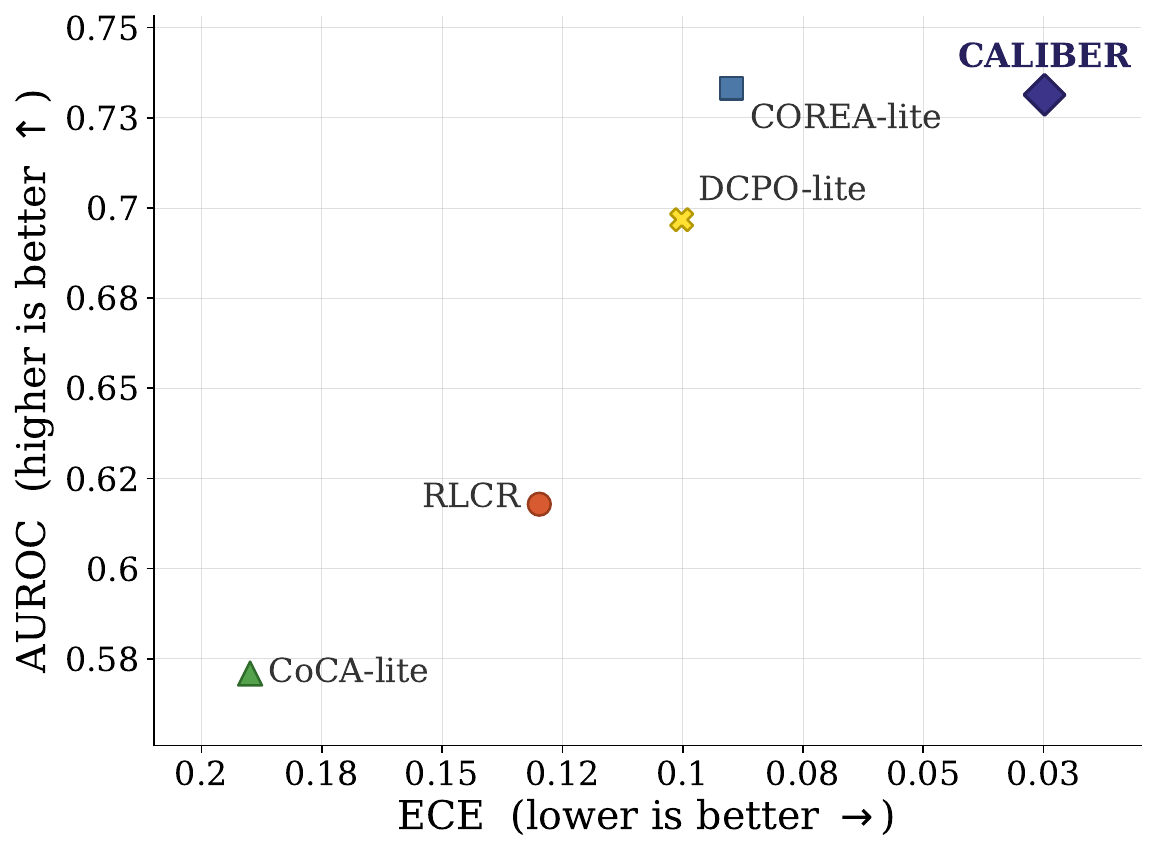}
        \caption{30B model.}
        \label{app:pareto_gpqa_30b}
    \end{subfigure}

    \caption{Pareto plots on GPQA.}
    \label{app:pareto_ood_gpqa}
\end{figure}

\begin{figure}[h]
    \centering
    \begin{subfigure}[t]{0.49\linewidth}
        \centering
        \includegraphics[width=\linewidth]{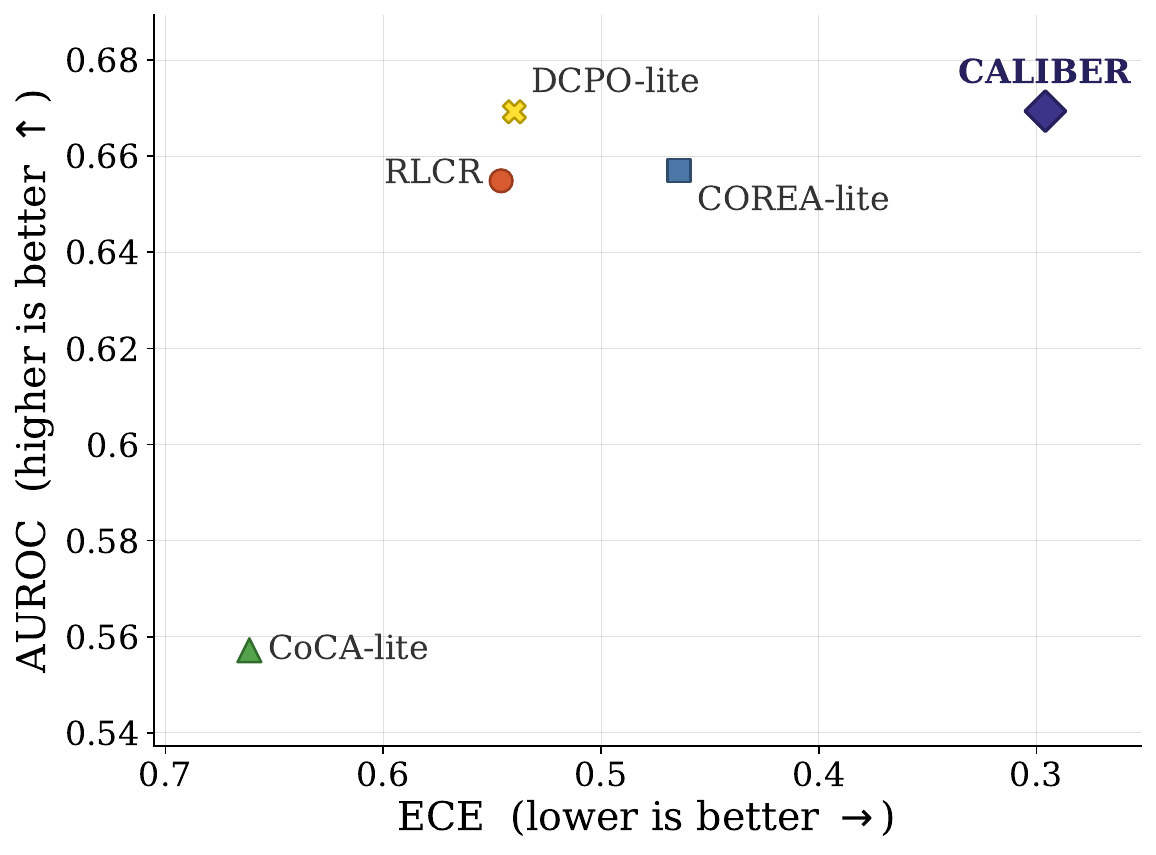}
        \caption{7B model.}
        \label{app:pareto_simpleqa_7b}
    \end{subfigure}
    \hfill
    \begin{subfigure}[t]{0.49\linewidth}
        \centering
        \includegraphics[width=\linewidth]{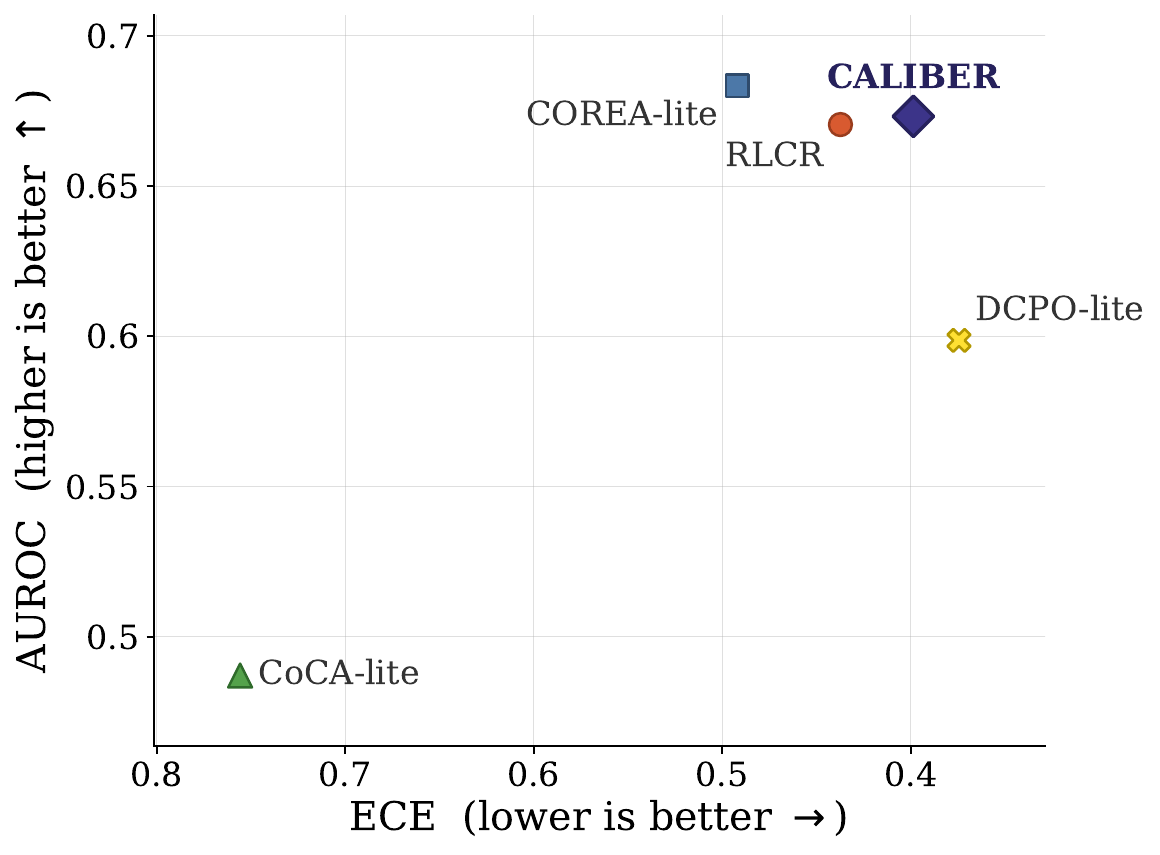}
        \caption{30B model.}
        \label{app:pareto_simpleqa_30b}
    \end{subfigure}

    \caption{Pareto plots on SimpleQA.}
    \label{app:pareto_ood_simpleqa}
\end{figure}

Figures~\ref{app:pareto_ood_triviaqa}--\ref{app:pareto_ood_simpleqa} show dataset-level ECE-AUROC tradeoffs for TriviaQA, GPQA and SimpleQA, complementing the average OOD Pareto plot in Figure~\ref{fig:ood-pareto}. Points farther to the right and higher on each plot correspond to lower calibration error and stronger failure prediction. Across datasets and model sizes, CALIBER generally lies on or near the favorable frontier: it achieves the lowest ECE on GPQA and TriviaQA for both model sizes, while retaining competitive AUROC, and gives the best 7B SimpleQA tradeoff. The main exception is 30B SimpleQA, where DCPO-lite obtains lower ECE, but CALIBER achieves substantially higher AUROC. These plots reinforce the main OOD conclusion that position-target alignment improves absolute calibration without collapsing confidence into an uninformative average, while also showing that low-accuracy settings such as SimpleQA expose a genuine calibration-discrimination tradeoff.

\end{document}